\documentclass[journal]{IEEEtran}
\usepackage{cite}
\usepackage[pdftex]{graphicx}
\usepackage{amsmath,amssymb,amsfonts}
\usepackage{balance}
\usepackage{array}
\usepackage{wrapfig}
\usepackage{multirow}
\usepackage{tabu}
\usepackage{csquotes}
\usepackage{xcolor,soul}
\usepackage{url}
\usepackage{float}
\usepackage{algorithm}
\usepackage{algorithmic}
\sethlcolor{yellow}

\ifCLASSOPTIONcompsoc
    \usepackage[caption=false, font=normalsize, labelfont=sf, textfont=sf]{subfig}
\else
\usepackage[caption=false, font=footnotesize]{subfig}
\fi

\newcommand{\RomanNumeralCaps}[1]{\MakeUppercase{\romannumeral #1}}

\def\BibTeX{{\rm B\kern-.05em{\sc i\kern-.025em b}\kern-.08em
    T\kern-.1667em\lower.7ex\hbox{E}\kern-.125emX}}

\begin{document}

\title{Multi-objective Digital Circuit Block Optimisation based on Cell Mapping in an Industrial EDA Flow}

\author{Linan~Cao,~\IEEEmembership{Member,~IEEE,}
        Simon~J.~Bale,
        and~Martin~A.~Trefzer,~\IEEEmembership{Senior~Member,~IEEE}
\thanks{The authors are with the Intelligent Systems and Nanoscience Research Group, Department of Electronic Engineering, University of York, York, North Yorkshire, YO10 5DD, United Kingdom. e-mail: (lc1492, simon.bale, martin.trefzer)@york.ac.uk}
}


\maketitle

\begin{abstract} 

Modern electronic design automation (EDA) tools can handle the complexity of state-of-the-art electronic systems by decomposing them into smaller blocks or cells, introducing different levels of abstraction and staged design flows. However, throughout each independent-optimised design step, overhead and inefficiency can accumulate in the resulting overall design. Performing design-specific optimisation from a more global viewpoint requires more time due to the larger search space, but has the potential to provide solutions with improved performance. In this work, a fully-automated, multi-objective (MO) EDA flow is introduced to address this issue. It specifically tunes drive strength mapping, preceding physical implementation, through multi-objective population-based search algorithms. Designs are evaluated with respect to their power, performance and area (PPA). The proposed approach is aimed at digital circuit optimisation at the block-level, where it is capable of expanding the design space and offers a set of trade-off solutions for different case-specific utilisation. We have applied the proposed MOEDA framework to ISCAS-85 and EPFL benchmark circuits using a commercial 65nm standard cell library. The experimental results demonstrate how the MOEDA flow enhances the solutions initially generated by the standard digital flow, and how simultaneously a significant improvement in PPA metrics is achieved.

\end{abstract}

\begin{IEEEkeywords}
Multi-objective, Evolutionary Computing, Design Space Exploration, Digital Flow, EDA.
\end{IEEEkeywords}

\IEEEpeerreviewmaketitle

\section{Introduction}
\IEEEPARstart{T}{he} process of building a digital integrated circuit using blocks or cells from a foundry is a common and mature approach in modern digital VLSI design. Comprehensive industry-standard EDA flows are available to tape out digital chips. Technology down-scaling enables high-density integrated circuits and the EDA tools therefore need to handle a large quantity of cells during the flow. To find possible optimal trade-off solutions in regard to PPA using appropriate library cells while consuming less turnaround time is the challenge of design optimisation~\cite{kahng2011vlsi}.

Standard cell libraries typically contain a large number of functions and each function has multiple cells differing in drive strength. This enables numerous possible combinations of logic functions or drive strengths depending on the design specifications and the required loads in circuit paths. The possible design space is thus huge and complex because a circuit might be composed of millions of gates. Different combinations of gates (drive strengths) thus can directly determine the PPA metrics of a circuit.

In addition, the parameter search space when building and optimising digital ICs will be further complicated with practical design rules and constraints in physical implementation. This can lead to the rise of optimisation difficulty that designs must meet multiple objectives simultaneously while satisfying all rules and constraints at the layout level, which might be beyond what experienced engineers can manually handle. Automatic efficient design space exploration approaches promise to balance multiple design objectives. Researchers both from academia and industry have focused on investigating design space in the synthesis, place and route flow or up to system level, and applying optimisation in the flow. A number of techniques have been adopted such as heuristic techniques~\cite{rao1993hierarchical}, machine learning~\cite{ma2019cross}\cite{kwon2019learning}, design-parameter tuning~\cite{anwar2016early,ziegler2016synthesis,kahng2018no}.

Population-based metaheuristic optimisation algorithms like multi-objective evolutionary algorithms (MOEAs) are widely-used existing techniques that can efficiently perform design space exploration and ultimately find a set of Pareto-optimised solutions. Many publications exist on applying evolutionary algorithms (EAs) or genetic algorithms (GAs) to VLSI design process from the system level down to the physical level, which also includes optimisation and design space exploration on individual design levels, such as standard cell library depletion~\cite{ricci2007evolutionary}, macro-cell placement optimisation~\cite{rahim2008performance}, gate-sizing-based soft error optimisation~\cite{sheng2009soft}, netlists partitioning~\cite{sait2006evolutionary}, circuit equivalence checking~\cite{vasicek2011global} and system-on-chip (SoC) design space optimisation~\cite{palesi2002multi}~\cite{ascia2002framework}.

However, limited research investigates how multi-objective optimisation techniques can fully integrate into industrial synthesis, place and route flows, and how well MOEAs can work in optimising designs down to physical layouts. An automated multi-objective optimisation flow crossing different design levels from a global perspective is required to recover performance which may otherwise be lost in generic overheads spread across the hierarchical design process.

This paper proposes a population-based evolutionary search to maintain the optimisation in multiple objectives through refining drive strengths of logic gates and applies it to a standard digital flow to enhance the design solution in the loop. Due to scaling behaviour of the problem domain and the optimisation algorithm, the proposed optimisation approach is best-suited for multi-objective design of IP/block-level circuits. The main contributions of this work are summarised as follows: 1) A multi-objective (MO) EDA optimisation framework, fully-compatible with an industrial digital flow from logic synthesis to physical implementation. 2) Global tuning of standard cell drive strength mapping using parameterised gate-level circuit netlists. 3) Enhanced trade-off design solutions with improved PPA metrics. 4) A methodology to seed the MOEA with a solution population across different circuit topologies for multi-objective design space optimisation. 5) Improved coverage of the feasible design space providing a set of Pareto-optimised solutions.

The paper is structured as follows: Section~\ref{section:related work} gives an overview of related work. Section~\ref{section:Flow} introduces the proposed MOEDA flow. Experiment setup is described in Section~\ref{section:Expt. setup}. Section~\ref{section:Experiments} presents the multi-objective optimisation results of each benchmark used. Section~\ref{section:DSE} presents the analysis of tool-generated design space and the multi-objective design space exploration based on it. Section~\ref{section:Conclusion} provides conclusions.

\section{Related Work}
\label{section:related work}
\subsection{Design Flow Modifications}
Modern digital integrated circuit (IC) design flow is a mature EDA process including various steps from register-transfer level (RTL) design, logic synthesis to physical implementation. As each step introduces its own level of abstraction (e.g. from cells to functions, from functions to blocks), any margin or error introduced will therefore accumulate and propagate. Hence, achieving a good solution in each step is crucial for the success of subsequent design steps and the quality of the overall solution. In addition, the abstraction introduced in each step may speed-up evaluation at the cost of optimal performance. Furthermore, standard cell libraries from a foundry do not allow transistor resizing or cell layout modifications when they are used in the digital flow. These limits may prevent EDA tools make full use of the capability of a process technology.

In previous work~\cite{cao2018instrumenting}, we introduced a customised multi-objective auto-design flow to adjust parameterised circuit layouts. This optimisation flow exploited a method to tune cell drive strengths using a scripted layout template, which aimed to achieve improved solutions in delay, energy and area.

Chinnery stated that there is a gap between full-custom design and standard digital flow in terms of speed and power~\cite{chinnery2005closing}~\cite{chinnery2000closing} in the 2000s. Digital ICs implemented using the standard design flow may significantly reduce design cycle time but have lost possible optimal trade-off solutions, which full-custom design can achieve. But designers in industry still focus on synthesis-centred methodology to save design efficiency due to the nowadays time-to-market pressure. 

Implementing extra custom design and optimisation techniques compensating to the standard digital flow can achieve better results~\cite{chinnery2013high}. Dally proposed to selectively apply a set of custom design methods in the digital flow, including custom floor-planning, place and route critical signals, to achieve the most compact layout structure~\cite{dally2000role}. To accelerate custom design,~\cite{onodera2001asic} introduced an ASIC design methodology with on-demand library generation in the digital flow producing cells with tailored drive strengths from a set of symbolic layouts. 

\subsection{Design Space Exploration using Standard Digital Flow}
Optimisation using steps of an industrial EDA flow in the loop can be viewed as black-box design space exploration. While many of the algorithms used in EDA flows are proprietary and not accessible by end users, logic synthesis and physical design tools provide a range of parameters and optimisation options for designers to choose from such as logic reconstruction, area constraints, synthesis effort level, place and route with timing or power optimisation, etc. These parameters can be tuned with an optimisation or machine learning approach to fully utilise the optimisation potential that the tools are capable of.

Kahng presented in~\cite{kahng2018no} that there is unpredictable~\enquote{noisiness} in tool-generated solutions causing variability in the resulting PPA metrics, and a probability theory was applied in a fully-automated digital flow, which aims to determine the optimal utilisation (parameter settings) of EDA tools to~\enquote{de-noise} the design results. In~\cite{ziegler2016synthesis}, an automated method to explore the search space via tuning parameters at the synthesis step for multi-objective optimisation in a rank-based iterative process is proposed.

Running through the whole design flow leads to more computing resource consumption. In~\cite{anwar2016early}, an automated selection mechanism based on searching the design space in parallel while pruning non-competitive solutions at early stage is exploited, rather than propagating through the entire design flow. In~\cite{ma2019cross}, machine learning (ML) approaches were employed to bridge the synthesis solution space to the physical solution space using a weighted sum cost function for solution evaluation, which aims to enable Pareto-driven exploration for high speed and power efficient adder designs. In~\cite{hyun2019accurate}, the authors propose ML-based methodologies to predict the actual wirelength of designs for better early-stage performance analysis. ML-based approaches are efficient to search a complex design space. However, ML techniques heavily rely on huge amounts of training data, which is not always readily available, possibly due to confidentiality in IC design area. Such issues are not present in the proposed MOEA. In EDA, the training process of MLs, which is compute and time intensive, also cannot be overlooked and puts requirements of other optimisation approaches in perspective.

\subsection{Discrete Gate Sizing for PPA Optimisation}
Gate sizing is a crucial step for achieving timing closure and power minimisation of ICs. It originally refers to determining transistor widths inside of logic gates to make designs meet constraints. Modern digital flows synthesize designs using a set of pre-defined cells. The optimisation problem thus is shifted to focusing on cell selection, regarding drive strengths and threshold voltage assignment, from discretised gate libraries.

A typical goal of gate sizing is to minimise power consumption while meeting timing requirements~\cite{lavagno2016electronic}. Lagrangian Relaxation (LR) is a recently adapted theory for gate sizing optimisation~\cite{flach2014effective,reimann2016cell,sharma2019lagrangian}, which moves the timing constraints to the objective function weighted by multipliers to penalise the overall results of the objective function. The problem is thus simplified to find the solution of weight factors.

In regard to the optimisation objectives in these LR-based approaches, ~\cite{flach2014effective} derived LR associated with finding trade-offs between leakage power and circuit timing.~\cite{reimann2016cell} expanded the primal objective function (power minimisation) by adding the area objective using an extra weight factor. More recently,~\cite{sharma2019lagrangian} considered more additional realistic constraints, such as maximum load, maximum slew, of gates for simultaneous gate sizing and clock skew scheduling. However, the Lagrangian relaxation, it is typically formulated for continuous problems and might not naturally handle discrete gate-sizing problems~\cite{hu2012sensitivity}~\cite{fatemi2019enhancing}.

There alternative multi-objective gate sizing frameworks, like geometric programming~\cite{farshidi2013self}~\cite{farshidi2014optimal}, simulated annealing~\cite{reimann2013simultaneous}, have been investigated using weighted sum objective functions which is a common scalarizing method in multi-objective problems similar to the LR.

In~\cite{hu2012sensitivity}, J. Hu proposed a different way to scalarise the objectives of leakage power and slacks into a sensitivity guided function for solution ranking (non-dominated), and a heuristic-based stochastic searching method was applied.
 
However, limited work completed the gate sizing with simultaneously handling all PPA metrics through industrial design flows and libraries, all the way from synthesis and physical implementation, to investigate how beneficial these methods can be in practice~\cite{fatemi2019enhancing}. The work in~\cite{yella2017standalone} stated that significant changes in cell sizes, after applying gate sizing optimisation, require re-placement and re-routing for new wire load parasitics. Therefore, optimising designs with timely updating corresponding layouts can make evaluations realistic, and achieved solutions feasible.

In earlier work, typical heuristic techniques like genetic algorithms were applied to solving gate sizing problems. The methods for multi-objective optimisation in~\cite{wang1995performance} and~\cite{benkhider2000parallel} both are still based on scalarized cost functions. More recently, gate-sizing-based soft error optimisation using MOEAs is proposed in~\cite{sheng2009soft} but its objectives are soft error rate, critical path delay and area. 

\subsection{Summary}
VLSI design is multi-objective in nature, often with a need to compromise between several conflicting design goals. A range of methods are developed including design flow revamping with custom-design techniques, intelligent approaches for design space exploration or dedicated design steps in EDA flows. In addition, the weighted sum function is often used in existing gate-sizing work as stated in previous literature. It is a popular linear scalarizing approach to decompose the complexity of multi-objective problems since its high search efficiency, and it is inherently used for convex problems~\cite{wang2016localized}. However, the physical characteristics of IC devices imply non-convexities and non-linearity~\cite{kahng2013high}, so the weighted sum method is not sufficient to search for feasible Pareto-optimal solutions~\cite{wang2016localized}.

Since solving the discrete gate-sizing problem still lacks theoretical guarantees~\cite{lavagno2016electronic} and still has been actively investigating, it is worthwhile to apply global search methods to optimise such a problem. Deterministic algorithms often used in EDA tools can always deliver the same solution for a given input with one execution, but might be limited to reach the possible global optimum. The MOEA, handling multiple design parameters and objectives inter-independently, is well-suited to perform the global search particularly regarding a large, complex design space.

\section{MOEDA Optimisation Framework}
\label{section:Flow}

\subsection{Preliminaries: Evolutionary Algorithms}
Evolutionary algorithms are a class of population-based metaheuristic optimisation algorithms using mechanisms inspired by biological evolution like reproduction, genetics and natural selection. An initial~\textit{population}, which consists of $N$ \textit{individuals} (candidate solutions), is allowed to age with~\textit{$M$} evolutionary~\textit{generations}. The $N$ is referred to the~\textit{population size}. The initial population can be either initialised randomly or seeded with a set of specific configurations. During each generation, individuals can be modified through~\textit{mutation} or \textit{crossover} (i.e., recombination with each other) variations on their~\textit{chromosomes}. All individuals are evaluated using a~\textit{fitness} function at the end of each generation. Only the fittest individuals survive the selection process for the subsequent generation. Termination of the evolution process is triggered when specific criteria are met like sufficient quality of solutions or maximum number of generations.

Applying an EA needs three main preparatory steps:

\textit{1) Definition of representation.} This is the data structure that the EA manipulates. It represents individuals as a set of genes (i.e., a~\textit{chromosome}) comprising all variables and parameters necessary to describe it.

\textit{2) Implementation of genetic operations.} Mutation and crossover are commonly applied in the evolution process. Mutation modifies genes of individuals, and crossover combines subsets of genes of multiple individuals to produce new ones.

\textit{3) Definition of a fitness function.} This is used to calculate a fitness score for each individual based on its performance regarding design objectives. The fitness scores are used during the ranking and selection process to determine which individuals survive to form the population for the next generation.

NSGA-\RomanNumeralCaps{2}~\cite{deb2002fast}, one of most popular MOEAs, has been adapted as the search tool in this work. The fast non-dominated sorting approach and diversity preservation strategies used ensure convergence while achieving a uniform spread of Pareto-optimal solutions.

\textit{Non-dominated sorting.} If one individual $p$ performs better than another $q$ in at least one objective while not performing worse in any other objectives, then $p$ is said to dominate $q$. In non-dominated sorting, each individual (e.g., $p$) has two entities: the first is domination count, the number of solutions that dominate $p$; the second is a set of solutions that $p$ dominates. The individuals are grouped based on their domination counts into multiple fronts $\mathbf{F} = (\mathbf{F}_{1},...,\mathbf{F}_{i})$. The non-dominated individuals which have the lowest domination counts (i.e., zero) form the first front $\mathbf{F}_{1}$. The individuals which have the second lowest domination counts form the second front $\mathbf{F}_{2}$ and this will continue to the third and following fronts until all individuals are assigned.

\textit{Diversity Preservation.} This~\textit{crowding distance sorting} algorithm estimates the solution density in the vicinity of each individual based on the Euclidean distance to their nearest neighbours. It mainly has two steps: the first is to calculate the distance of each individual to others, and assign the value to each individual; the second is to decendingly re-sort (\textit{Descend-Sort}) the $\mathbf{F}$ according to their distance values. So that if two individuals belong to the same non-dominated front, the one that resides in the less crowded region is preferred.

\begin{figure}[t]
    \centering
    \includegraphics[width=0.65\linewidth]{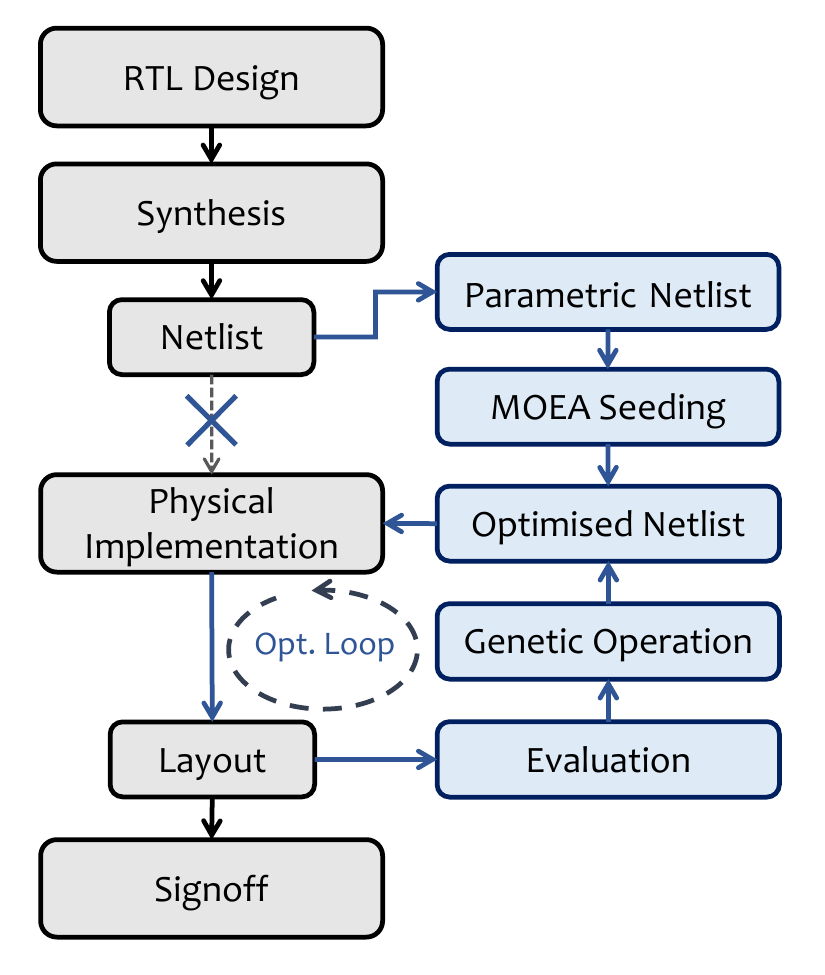}
    \caption{MOEDA Digital Flow. The flowchart on the left side is the standard digital flow and on the right side the MO extension is shown.}
    \label{fig:MOO_flow}
\end{figure}

\subsection{Multi-objective (MO) EDA Digital Flow}
The MOEDA digital flow, illustrated in Fig.~\ref{fig:MOO_flow}, is a fully-automated multi-objective design framework using compatible with an industrial digital flow. The industrial flow is tapped between the logic synthesis and the physical implementation stage, where the MO evolutionary optimisation loop is inserted. The novelty here lies in the additional level of abstraction that can automatically fine-tune drive strength mapping during the process of the flow. The proposed flow involves:

\textit{1) Parametric netlist.} A synthesised netlist is composed of technology-specified logic gates and their connectivity. The MOEA representation encodes the drive strengths of gates into a set of genes, a string vector $\mathbf{g}$ (i.e., instance names), defining each gate function and its drive strength. This information is used to produce a parametric netlist from the synthesis results.

\textit{2) MOEA seeding.} In this work, initial populations are seeded from the solutions obtained from the synthesis tool. This is achieved by converting the output netlists from the standard tool to parametric netlists, allowing the MOEA to optimise them. 

\textit{3) Genetic operations.} Only mutation operator is used in this work. The mutation operation modifies the drive strength of components based on a given probability $\rho$ (i.e., determining how many components out of all will be modified). This results in a new netlist, which is then ready for physical implementation. With the pressure to promote beneficial mutations and discard the others, the evolutionary loop continues to keep producing increasingly optimised solutions.

\textit{4) Evaluation.} This calculates the fitness scores of each individual. MOEA-optimised netlists are propagated to the physical implementation step, producing layout instances for accurate evaluation metrics. Three objectives are used here which are worst case delay ($D_{wc}$), total consumption power ($P_{total}$) and area of all logic gates ($A_{gate}$), and fitness scores are evaluated at post-route stage from the place and route tool. Fitness scores are then fed back to the MOEA for ranking and selection. 

The optimisation goal in this work is to simultaneously minimise $D_{wc}$, $P_{total}$ and $A_{gate}$ so the fitness function is:
\begin{equation}
\begin{aligned}
\displaystyle f(\mathbf{g}) = \min \quad [D_{wc}(\mathbf{g}), \quad P_{total}(\mathbf{g}), \quad A_{gate}(\mathbf{g})]\\
\textrm{s.t.} \quad  \mathbf{g}=(g_1, g_2, ..., g_i), \quad 3 \leq g_{i} \leq 11, \quad \forall g_{i} \in \mathbb{G}
\end{aligned}
\end{equation}
where the chromosome vector $\mathbf{g}$ is the input variables to the fitness function, which are drive strengths of gates ($g_{i}$) selected from a standard cell library ($\mathbb{G}$). There are between 3 and 11 drive options for each logic gate in $\mathbb{G}$. Hence, the average synthesised design space size of, for example, the log2 circuit (used in the experiment) is between $3^{10,000}$ and $11^{10,000}$.

\begin{figure}[t]
    \centering
    \includegraphics[width=\linewidth]{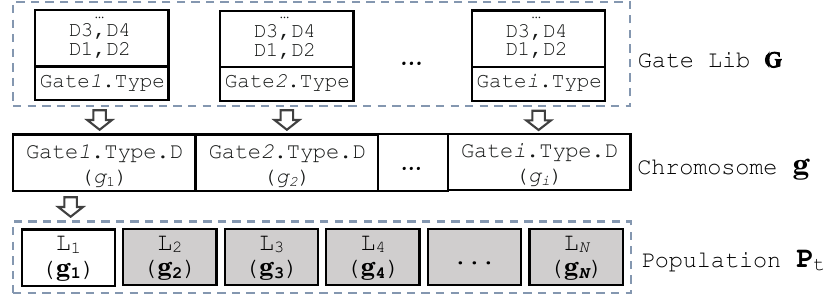}
    \caption{A chromosome example of an individual in a population and how each gene is mutated using a logic gate library. ``D'' represents the drive strength.}
    \label{fig:chrom_example}
\end{figure}

Fig.~\ref{fig:chrom_example} demonstrates a population example where $\mathbf{P}_{t}$ consists of $N$ layout individuals ($\mathrm{L_{1}, L_{2}, ..., L_{n}}$). Each $\mathrm{L}$ has a chromosome $\mathbf{g}$ comprising a set of genes ($g_1, g_2, ..., g_i$). The chromosome overall represents the all logic gates of a netlist. Each single $g$ ($\mathrm{Gate.Type.D}$) represents a logic gate ($\mathrm{Gate}$) including its properties: function type ($\mathrm{Type}$) and drive strength size ($\mathrm{D}$). When mutation is triggered, the gates to be mutated are randomly selected according to the mutation rate $\rho$. For each selected gate, it will firstly identify its function ($\mathrm{Gate.Type}$) and then perform an online look-up to achieve the all drive strength options ($\mathrm{D}$) of this gate function from $\mathbb{G}$, and finally choose one from them to replace the previous one.

The overall optimisation process, presented in Algorithm 1, is continuously producing different circuit layout instances by adjusting the netlists and keeping improved solutions generation-by-generation.

\begin{algorithm}[t]
\caption{Adapted NSGA-\RomanNumeralCaps{2} for MOEDA~\cite{deb2002fast}}
\textbf{Procedure:} NSGA-\RomanNumeralCaps{2} ($N$, $M$, $f(\mathbf{g})$) $\triangleright$ $N$ individuals evolved $M$ generations to solve $f(\mathbf{g})$.
\begin{algorithmic}[1]
\STATE Initialize parent population $\mathbf{P}_{t}$ in size $N$ $\triangleright$ Seed with synthesis-optimised solutions generated by the tool.
\STATE Offspring population $\mathbf{Q}_{t}$ $\gets$ Mutation($\mathbf{P}_{t}$)
\FOR{$t \gets 1$ to $M$}
\FOR{each population $\mathbf{R}_{t} \gets \mathbf{P}_{t} \cup \mathbf{Q}_{t}$ in size $2N$}
\STATE Fitness evaluation $\gets$ $f(\mathbf{\mathbf{R}_{t}})$ $\triangleright$ Call fitness function $f(\mathbf{g})$ for each individual evaluation.
\STATE $\mathbf{F}$ $\gets$ Non-Dominated-Sorting($\mathbf{R}_{t}$)
\STATE $\mathbf{P}_{t+1} \gets$ \O
\STATE $i \gets 1$
\WHILE{$|\mathbf{P}_{t+1}| + |\mathbf{F}_{i}| \leq N$}
\STATE Crowding-Distance-Assignment($\mathbf{F}_{i}$)
\STATE $\mathbf{P}_{t+1}$ $\gets$ $\mathbf{P}_{t+1} \cup \mathbf{F}_{i}$
\STATE $i \gets i + 1$
\ENDWHILE
\STATE $\mathbf{F}_{i} \gets$ Descend-Sort($\mathbf{F}_{i}$)
\STATE $\mathbf{P}_{t+1} \gets \mathbf{P}_{t+1} \cup \mathbf{F}_{i}[1:(N-|\mathbf{P}_{t+1}|)]$ $\triangleright$ Less crowded individuals from the first to the $(N-|\mathbf{P}_{t+1}|)$th of $\mathbf{F}_i$ to fill $\mathbf{P}_{t+1}$.
\STATE $\mathbf{Q}_{t+1} \gets$ Mutation($\mathbf{P}_{t+1}$)
\ENDFOR
\ENDFOR
\end{algorithmic}
\end{algorithm}

\section{Experiment Setup}
\label{section:Expt. setup}
We implement the proposed algorithm in C++ and conduct the proposed MOEDA design flow experiments on a 2.2GHz Xeon E5-2650 CPU. The benchmark circuits from ISCAS-85~\cite{iscas} and EPFL~\cite{amaru2015epfl} are implemented and optimised using the Cadence\textsuperscript{\textregistered} digital flow suite. Benchmark circuits in the form of RTL designs are synthesised into gate-level netlists using Genus\textsuperscript{TM} (v17.11). These netlists are then optimised using the proposed flow in tandem with the physical implementation tool Innovus\textsuperscript{TM} (v17.11) to generate the layouts from the optimised netlists. The versions of used EDA tools represented the most up-to-date flow when we performed the experiments. We also have full optimisation licences of Cadence\textsuperscript{\textregistered} digital flow. All experiments are using a TSMC 65nm low-power core cell library (TCBN65LP) in standard threshold voltage.


\subsection{Tool Environment Setup}
The MOEDA flow is applied to further enhance designs which are already well-optimised by the Cadence\textsuperscript{\textregistered} tools. In order to take advantage of the Genus\textsuperscript{TM} synthesis tool as much as possible, it is necessary to push it to the limit of what it can achieve with the user options available. Hence, the \textit{synthesis compile effort} is set to high and \textit{ultra optimisation} is enabled. Apart from that, each benchmark is repeatedly synthesised, tightening its timing constraint bit-by-bit until it fails timing. The last working solution before timing failure is the best in speed, delay or slack that the tool can achieve. This solution is then chosen as a seed for initialising the MOEA. 

\begin{figure}[t]
    \centering
    \includegraphics[width=0.9\linewidth]{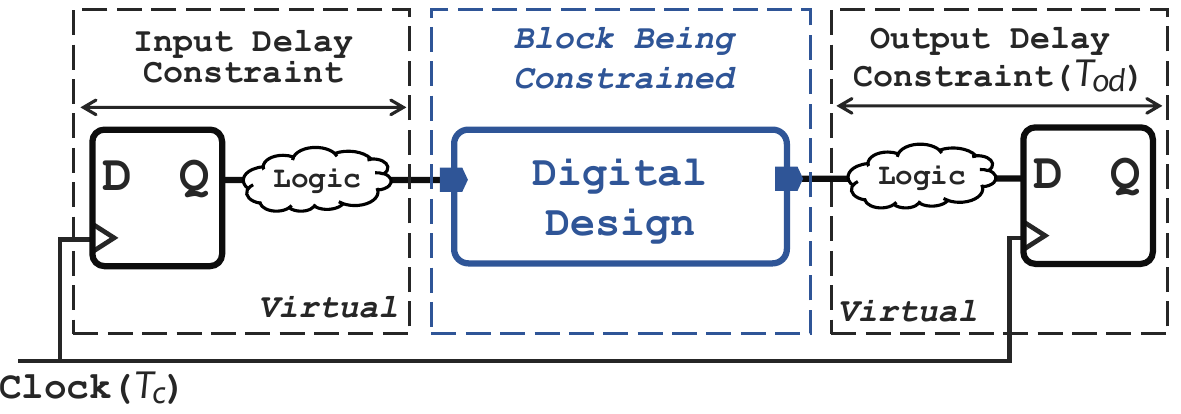}
    \caption{Conceptual testbench to define the timing constraints in the EDA tools. Input/output delay constraints are set by using virtual logic parts and flip-flops, allowing users to specify clocks and timing requirements. The~\textit{Digital Design} in the middle is constrained.}
    \label{fig:timing_constraints}
\end{figure}

In the timing constraint setup, we create an ideal general clock for all inputs and outputs, which means all paths are clocked with two ideal flip-flops at the beginning and the end of each path shown in Fig.~\ref{fig:timing_constraints}. The benchmarks used are all combinational circuits, so that the ideal clock was not applied with any uncertainties or transition delays.

To tighten the timing constraint, shown in Fig.~\ref{fig:timing_constraints}, the output delay constraint ($T_{od}$) is gradually increased for a given clock period ($T_{c}$). The required time ($T_{r}$) is calculated through Equation~\eqref{eq:required_time} in EDA tools, so the required time ($T_{r}$) that solutions need to meet is gradually becoming strict.
\begin{equation}
    T_{r} = T_{c} - T_{od}
\label{eq:required_time}
\end{equation}

\begin{table}[t]
\begin{center}
\caption{Tool Settings for Synthesis and Physical Implementation}
\begin{tabular}{|c|c|}
  \hline
  Synthesis Setup & Place \& Route Setup\\ 
  \hline
  \multirow{3}{*}{syn\_generic\_effort = high}      &die aspect ratio = 1.0 \\
  \multirow{3}{*}{iopt\_ultra\_optimisation = true} &core utilisation = 0.7 \\
                                                    &timing-driven placement = true \\
                                                    &timing-driven routing = true \\
                                                    &SI-driven routing = true \\
  \hline
\end{tabular}
\label{table:tool_setting}
\end{center}
\end{table}

The circuit path arrival time ($T_{a}$) should be less than the required time ($T_{r}$) to the meet the timing constraint. The settings of both synthesis step and physical implementation step are summarised in Table~\ref{table:tool_setting}.

The output load capacitance (set\_load) is also specified in part of the following experiments. In the physical design flow, all die area is shaped in the ratio of 1.0, and core utilisation is 70\%. Timing-driven placement and routing, and signal integrity (SI) driven routing are enabled for better performance.

\subsection{Objective Evaluation in Tools}
In this work, the evaluation regarding three objectives takes place after place-and-route with Innovus\textsuperscript{TM} as follows:

\textit{$D_{wc}$}, worst case $T_{a}$ which is the value of $T_{r}$ minus the worst negative slack amongst all path delays. This is achieved by performing static timing analysis at the post-route stage.

\textit{$P_{total}$}, which is the results reported by power analysis in Innovus\textsuperscript{TM}. It includes switching power, internal power (short-circuit power) and leakage power. Both internal and leakage power are calculated based on power tables provided in the Liberty (.lib) file, which contains the specifications and characterisations of the standard cells. Switching power is calculated based on the equation $\mathrm{P = 0.5 \ast C_{L}V^{2}F \ast A}$, where $\mathrm{C_{L}}$ is the output capacitive loading, $\mathrm{V}$ is the voltage, $\mathrm{F}$ is frequency, and $\mathrm{A}$ is the average switching activity (the value 0.2 used in this work is the default from Innovus\textsuperscript{TM}).

\textit{$A_{gate}$}, which is calculated by adding the areas of each single gate used. This is directly reported by Innovus\textsuperscript{TM}.

\begin{table}[htbp]
\begin{center}
\caption{Statistics of Benchmarks}
\begin{tabular}{|c|c|c|c|}
  \hline
  Test Case  &No. Inputs &No. Outputs &No. Gates \\\hline
  C1908 &33  &25  &880   \\\hline
  C5315 &178 &123 &2307  \\\hline
  C6288 &32  &32  &2406  \\\hline
  log2  &32  &32  &32060 \\
  \hline
\end{tabular}
\label{table:benchmarks}
\end{center}
\end{table}

All evaluations above are performed on a single mode under typical corner conditions (PVT: TT, 1.2V, 25$^{\circ}$C).

\begin{table*}[ht]
\centering
\caption{MOEDA design flow with using the full commercial library}
\begin{tabular}{|c|c|c|c|l|c|c|c|}
\multicolumn{4}{l}{Units: $D_{wc}$ [$ns$] \space\space $P_{total}$ [$uW$] \space\space $A_{gate} $[$um\textsuperscript{2}$]} & \multicolumn{4}{r}{$N=200$, $M=200$, $\rho=1\%$, set\_load=0} \\
    \hline
     Test & clock   &\multirow{2}{*}{(No.) $T_{r}$} &No. Syn Gates &\multicolumn{1}{c|}{Syn-Opt.} &\multicolumn{3}{c|}{MOEDA Solution} \\\cline{6-8}
     Case &($T_{c}$)&                               &No. Genes     &\multicolumn{1}{c|}{Solution} &Best $D_{wc}$ ($\Delta\%$) &Best $P_{total}$ ($\Delta\%$) &Best $A_{gate}$ ($\Delta\%$) \\
    \hline	
    \multirow{9}{*}{C1908}&\multirow{9}{*}{250MHz} &\multirow{3}{*}{$(a)$ 0.60$ns$} &\multirow{2}{*}{299} &$D_{wc}$:    \textbf{0.580}  &\textbf{0.569 (1.9\%)}&0.580                 &0.580                   \\
                          &                        &                                &\multirow{2}{*}{299} &$P_{total}$: \textbf{222.9}  &221.9                 &\textbf{211.0 (5.3\%)}&211.0                   \\
                          &                        &                                &                     &$A_{gate}$:  \textbf{1452.96}&1451.88	           &1388.16               &\textbf{1388.16 (4.5\%)}\\\cline{3-8}
                          
                          &                        &\multirow{3}{*}{$(b)$ 0.76$ns$} &\multirow{2}{*}{178} &$D_{wc}$:    \textbf{0.697}  &\textbf{0.687 (1.4\%)}&0.688                 &0.696                  \\
                          &                        &                                &\multirow{2}{*}{178} &$P_{total}$: \textbf{111.1}  &107.9                 &\textbf{107.5 (3.2\%)}&0.1078                 \\
                          &                        &                                &                     &$A_{gate}$:  \textbf{698.04} &682.92                &682.2                 &\textbf{678.96 (2.7\%)}\\\cline{3-8}
                           
                          &                        &\multirow{3}{*}{$(c)$ 1.50$ns$} &\multirow{2}{*}{105} &$D_{wc}$:    \textbf{1.263}  &\textbf{1.234 (2.3\%)}&1.249                 &1.251                  \\
                          &                        &                                &\multirow{2}{*}{105} &$P_{total}$: \textbf{42.1}   &39.69                 &\textbf{39.32 (6.6\%)}&39.51	              \\
                          &                        &                                &                     &$A_{gate}$:  \textbf{344.52} &344.52	               &343.08                &\textbf{342.72 (0.5\%)}\\
    \hline
    \multirow{9}{*}{C5315}&\multirow{9}{*}{250MHz}  &\multirow{3}{*}{$(a)$ 0.74$ns$} &\multirow{2}{*}{750} &$D_{wc}$:    \textbf{0.723}   &\textbf{0.706 (2.4\%)}&0.715                 &0.72                    \\
                          &                         &                                &\multirow{2}{*}{750} &$P_{total}$: \textbf{472.9}   &470.5                 &\textbf{458.9 (3.0\%)}&461.2                   \\
                          &                         &                                &                     &$A_{gate}$:  \textbf{2762.64} &2755.44               &2729.16               &\textbf{2724.48 (1.4\%)}\\\cline{3-8}
                           
                          &                         &\multirow{3}{*}{$(b)$ 0.88$ns$} &\multirow{2}{*}{516} &$D_{wc}$:    \textbf{0.824}   &\textbf{0.805 (2.3\%)}&0.819                 &0.823                   \\
                          &                         &                                &\multirow{2}{*}{516} &$P_{total}$: \textbf{310.9}   &309.0                 &\textbf{304.6 (2.0\%)}&305.7                   \\
                          &                         &                                &                     &$A_{gate}$:  \textbf{1873.44} &1869.48               &1859.76               &\textbf{1852.56 (1.1\%)}\\\cline{3-8}

                          &                         &\multirow{3}{*}{$(c)$ 1.50$ns$} &\multirow{2}{*}{400} &$D_{wc}$:    \textbf{1.305}   &\textbf{1.241 (4.9\%)}&1.289                 &1.302                   \\
                          &                         &                                &\multirow{2}{*}{400} &$P_{total}$: \textbf{225.2}   &222.3                 &\textbf{217.4 (3.5\%)}&220                     \\
                          &                         &                                &                     &$A_{gate}$:  \textbf{1346.76} &1343.52               &1343.16               &\textbf{1336.68 (0.8\%)}\\
    \hline		
    \multirow{9}{*}{C6288}&\multirow{9}{*}{250MHz}  &\multirow{3}{*}{$(a)$ 2.34$ns$} &\multirow{2}{*}{2178} &$D_{wc}$:    \textbf{2.225}   &\textbf{2.204 (0.9\%)}&2.206                &2.204                   \\
                          &                         &                                &\multirow{2}{*}{2178} &$P_{total}$: \textbf{5509}    &5495                  &\textbf{5481 (0.5\%)}&5495                    \\
                          &                         &                                &                      &$A_{gate}$:  \textbf{9382.32} &9364.68               &9377.28              &\textbf{9364.68 (0.2\%)}\\\cline{3-8}
                           
                          &                         &\multirow{3}{*}{$(b)$ 2.90$ns$} &\multirow{2}{*}{1555} &$D_{wc}$:    \textbf{2.726}   &\textbf{2.673 (1.9\%)}&2.708                &2.708                   \\
                          &                         &                                &\multirow{2}{*}{1555} &$P_{total}$: \textbf{3829}    &3785                  &\textbf{3732 (2.5\%)}&3732                    \\
                          &                         &                                &                      &$A_{gate}$:  \textbf{6363.00} &6331.32               &6278.76              &\textbf{6278.76 (1.3\%)}\\\cline{3-8}
                           
                          &                         &\multirow{3}{*}{$(c)$ 4.00$ns$} &\multirow{2}{*}{1140} &$D_{wc}$:    \textbf{3.591}   &\textbf{3.528 (1.8\%)}&3.59                 &3.585                   \\
                          &                         &                                &\multirow{2}{*}{1140} &$P_{total}$: \textbf{2824}    &2821                  &\textbf{2754 (2.5\%)}&2777	                 \\
                          &                         &                                &                      &$A_{gate}$:  \textbf{4194.00} &4191.84               &4183.92              &\textbf{4137.48 (1.3\%)}\\
    \hline
    \multirow{9}{*}{log2}&\multirow{9}{*}{40MHz}   &\multirow{3}{*}{$(a)$ 16.4$ns$} &\multirow{2}{*}{11838} &$D_{wc}$:    \textbf{16.355}   &\textbf{15.839 (3.2\%)}&16.24                 &16.355                \\
                         &                         &                                &\multirow{2}{*}{11838} &$P_{total}$: \textbf{19610}    &19090                  &\textbf{19070 (2.8\%)}&19610                 \\
                         &                         &                                &                       &$A_{gate}$:  \textbf{38547.0}  &38728.1 (-0.5\%)       &38797.6 (-0.6\%)      &\textbf{38547.0 (0.0\%)}\\\cline{3-8}

                         &                         &\multirow{3}{*}{$(b)$ 17.9$ns$} &\multirow{2}{*}{11272} &$D_{wc}$:    \textbf{17.751}   &\textbf{17.364(2.2\%)}&17.72                 &17.751                  \\
                         &                         &                                &\multirow{2}{*}{11272} &$P_{total}$: \textbf{18000}    &18000                 &\textbf{17890 (0.6\%)}&18000                   \\
                         &                         &                                &                       &$A_{gate}$:  \textbf{36623.9}  &36840.6 (-0.6\%)      &36726.8 (-0.3\%)      &\textbf{36623.9 (0.0\%)}\\\cline{3-8}

                         &                         &\multirow{3}{*}{$(c)$ 18.8$ns$} &\multirow{2}{*}{11119} &$D_{wc}$:    \textbf{18.435}   &\textbf{17.795 (3.5\%)}&18.357                &18.435                  \\
                         &                         &                                &\multirow{2}{*}{11119} &$P_{total}$: \textbf{17590}    &17510                  &\textbf{17390 (1.1\%)}&17590	                \\
                         &                         &                                &                       &$A_{gate}$:  \textbf{35999.6}  &36227.5 (-0.6\%)	    &36203.0 (-0.5\%)      &\textbf{35999.6 (0.0\%)}\\
    \hline
\end{tabular}
\label{table:iscas_full}
\end{table*}

\subsection{Multi-threads Running and Runtime}
According to the computing resources and licenses, we run all experiments in this work parallelly using 24 threads in an MOEDA run for evaluating individuals.

The multi-objective approach requires a larger number of evaluations, which increases the runtime of the algorithm. The majority of runtime is spent on completing place and route in this case. This aims to achieve accurate metrics as close as possible to sign-off. Therefore, the runtime is not the primary concern in this work.

However, due to the inherent parallelism of the population-based approach, this can be overcome using a larger number of licenses and high-performance computing (HPC) resources. In addition, the MOEDA algorithm feature is able to deliver a set of trade-off solutions spanning the feasible design space in one go, rather than a single, case-specific solution. In~\cite{hyun2019accurate}, the authors proposed ML techniques to improve the speed and accuracy of multi-objective design space exploration problems. We also see the most powerful solutions in the future when combining different methods appropriately at different levels and stages of the design hierarchy, i.e., ML + EA, but this is not the focus of this work.

\section{Multi-objective Optimisation Experiments}
\label{section:Experiments}

\subsection{Experiments with a Full Foundry Library}
\label{section:exp:fulllib}

In this set of experiments, the selected three benchmarks from ISCAS-85 suite, in different structures and functions, are a 16-bit error detector/corrector (C1908), a 9-bit ALU (C5315) and a 16x16 multiplier (C6228). One large circuit used from EPFL benchmark suite is an arithmetic function for log2 calculation. The statistics of benchmarks are summarised in Table~\ref{table:benchmarks}. The reason that we only use combinational circuits is all large squential circuits are built from basic combinational blocks, and the optimisation of a sequential circuit will eventually collapse into the optimisation of its combinational parts~\cite{amaru2015epfl}. In addition, although most gate-sizing related research optimises sequential circuits, they still only manipulate on combinational components~\cite{flach2014effective}\cite{hu2012sensitivity,sharma2019lagrangian,fatemi2019enhancing}.

\begin{figure*}[ht]
  \centering
     \subfloat{\includegraphics[width=0.25\linewidth]{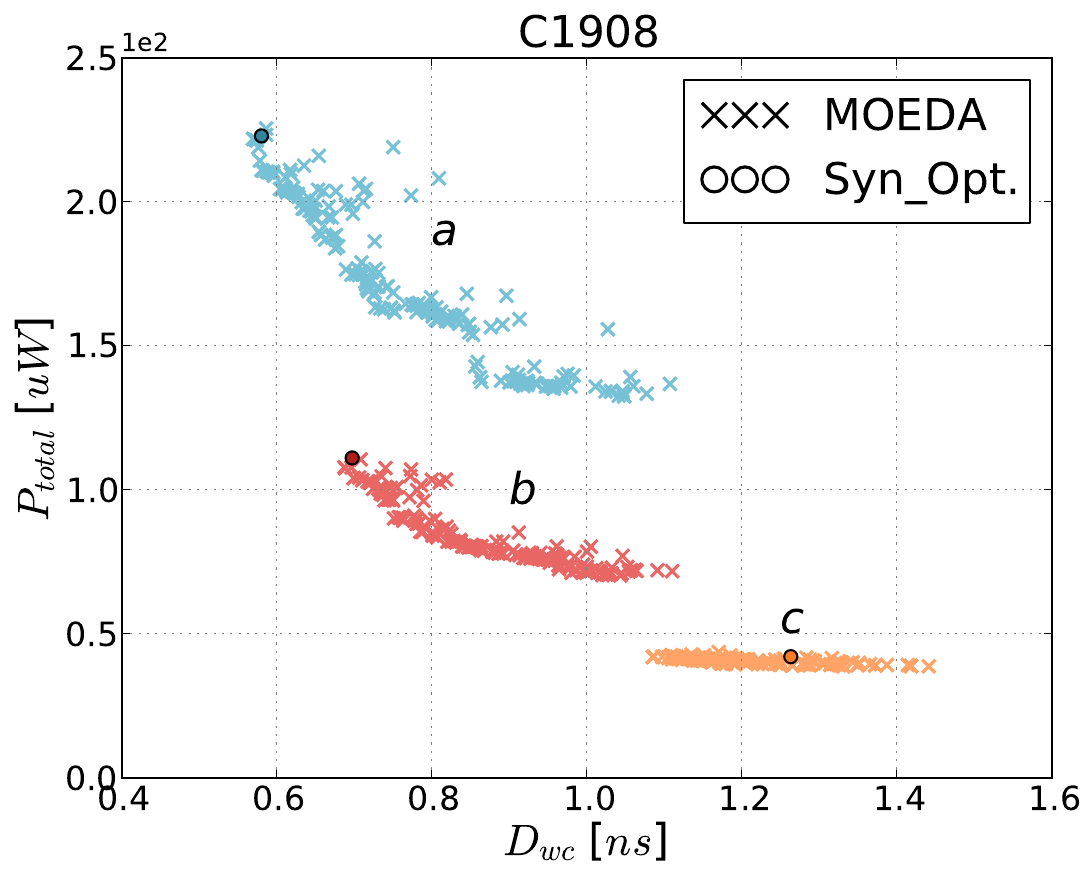}}
     \subfloat{\includegraphics[width=0.25\linewidth]{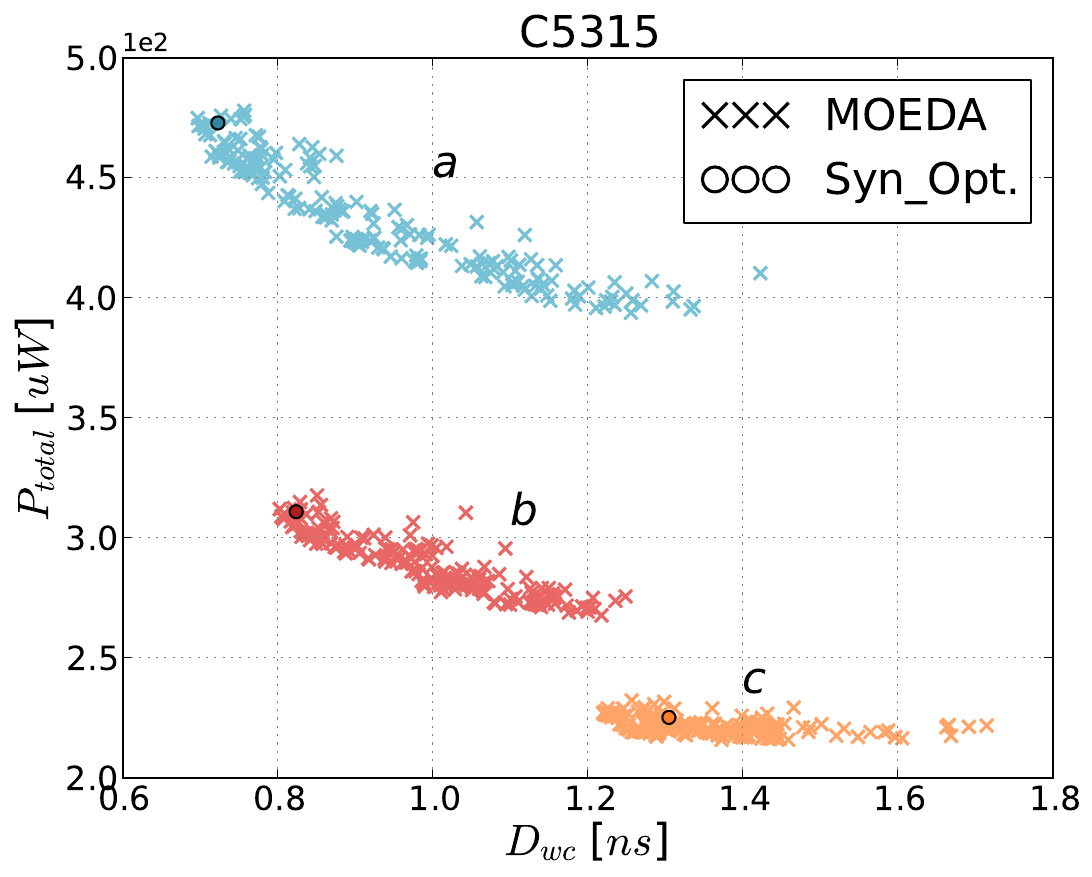}}
     \subfloat{\includegraphics[width=0.25\linewidth]{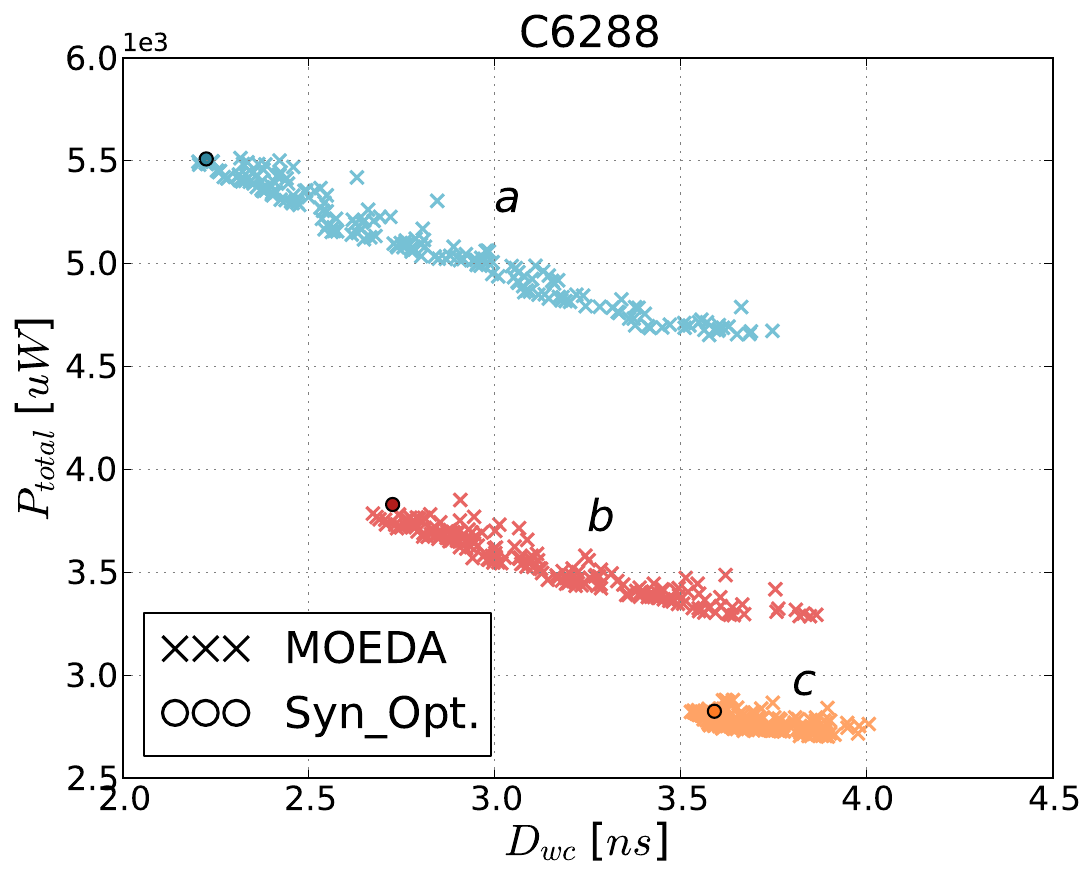}}
     \subfloat{\includegraphics[width=0.25\linewidth]{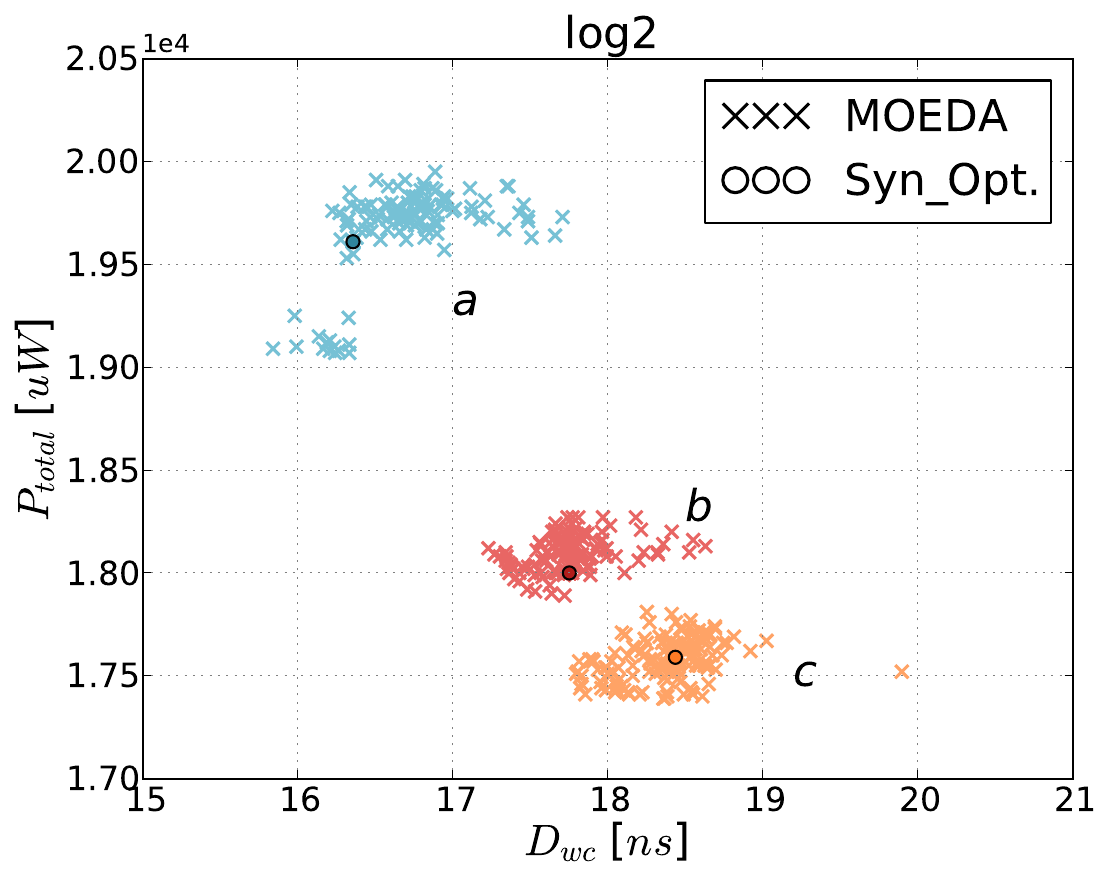}}\\
     \subfloat{\includegraphics[width=0.25\linewidth]{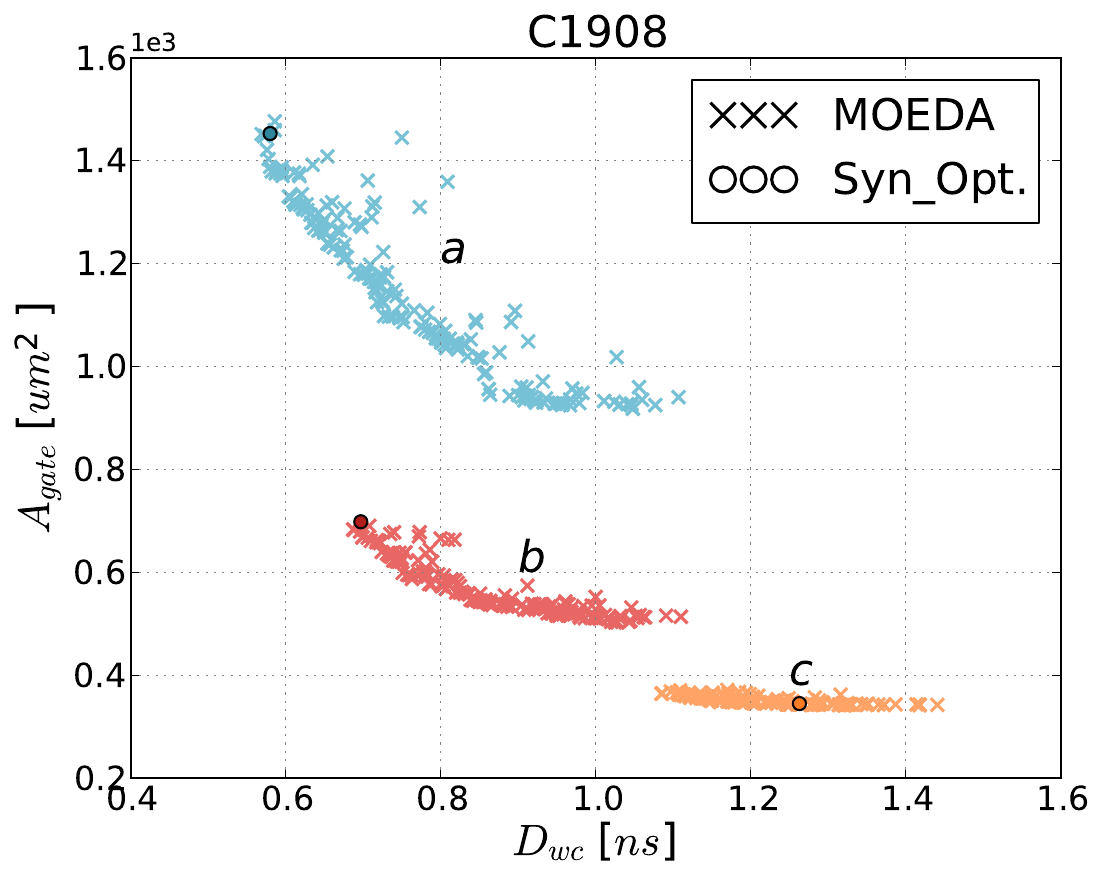}}
     \subfloat{\includegraphics[width=0.25\linewidth]{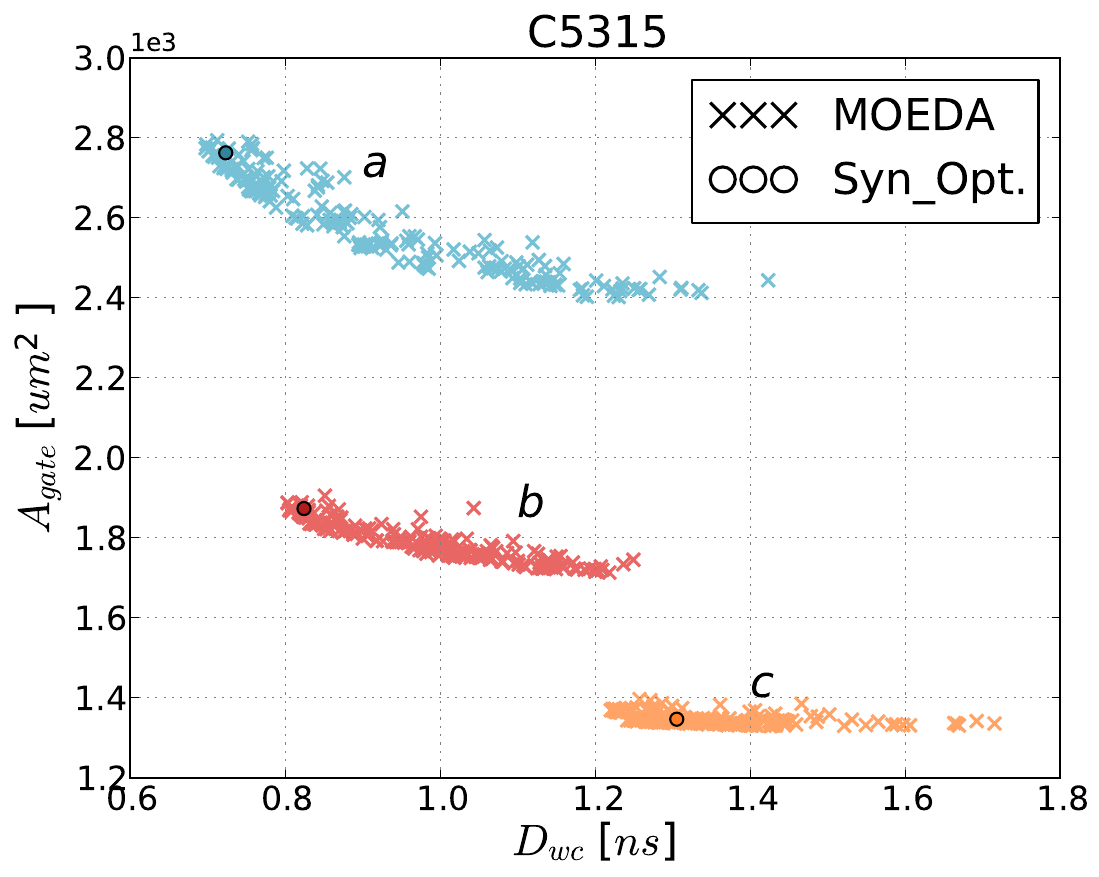}}
     \subfloat{\includegraphics[width=0.25\linewidth]{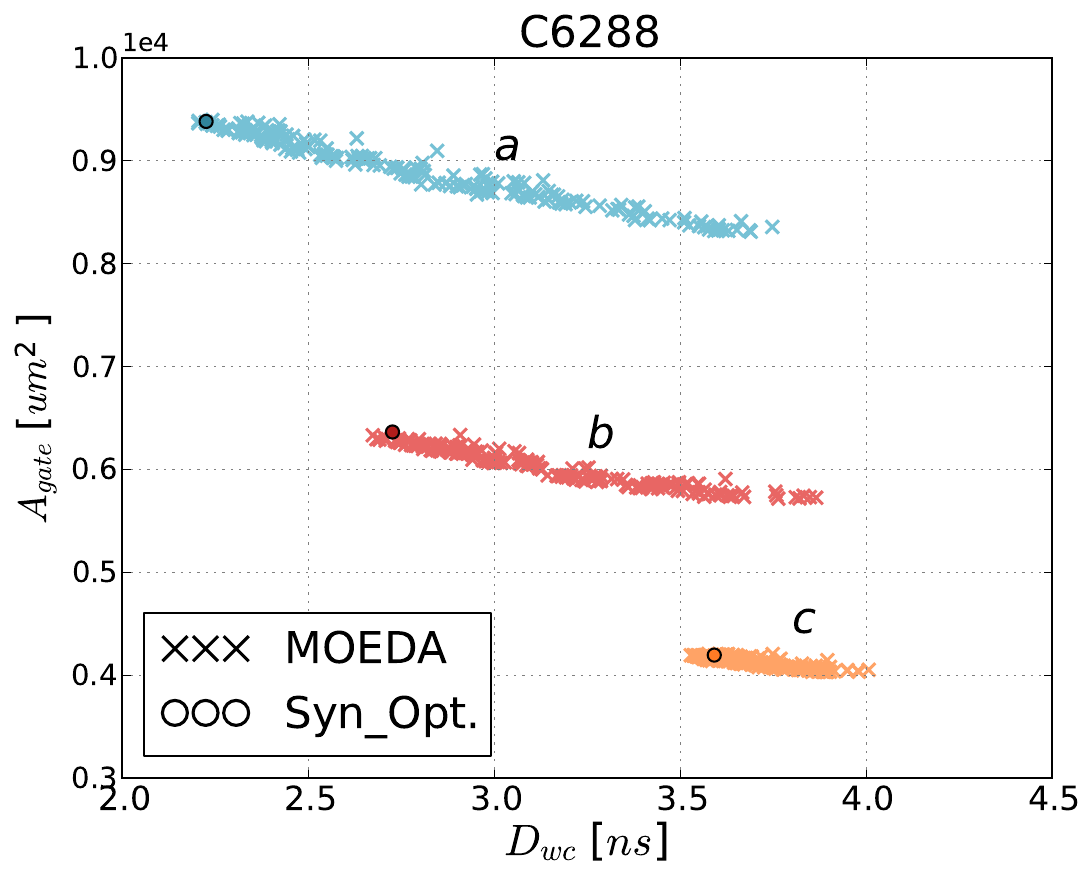}}
     \subfloat{\includegraphics[width=0.25\linewidth]{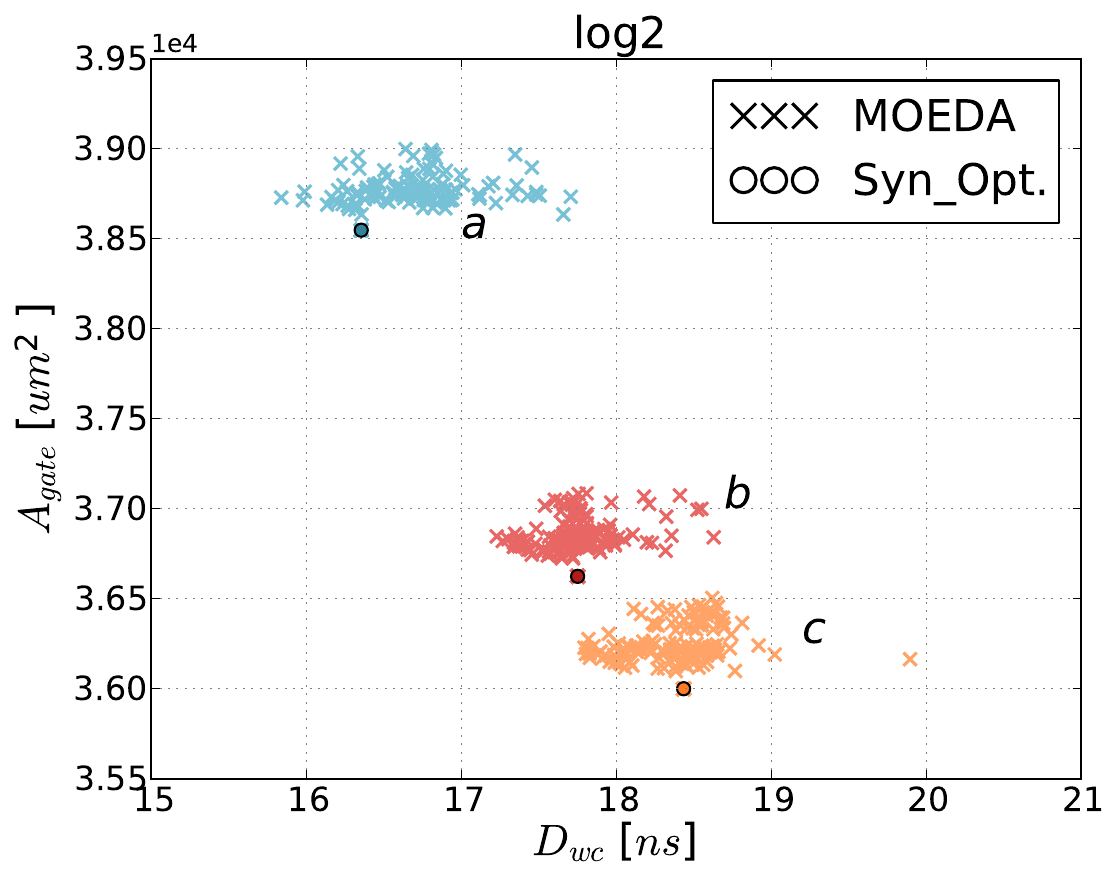}}
     \caption{MOEDA flow optimisation results using full commercial standard cell library for C1908, C5315, C6288 and log2 circuits.}
     \label{fig:MOO_Full}
\end{figure*}

The proposed MOEDA flow can handle the drive strength optimisation for all types of cells available from the TSMC TCBN65LP library. Three different seeds are used here to initialise the MOEDA algorithm, which are obtained from running synthesis and implementation under three different timing constraints for each benchmark: the first (named $a$ later) is the tightest constraint that can just be met, resulting in a solution with the best delay. In the second case, the timing constraint is relaxed so that it can be easily met, allowing the standard flow room to optimise for power and area. The third timing constraint is chosen in the middle of the first and second. The three different solutions obtained will be used as seeds to perform three independent runs of the MOEDA flow. This aims to investigate how the synthesis tool optimises solutions in trading off PPA metrics when setting different timing goals, and how the MOEDA flow further compensates these tool-generated solutions.

In this work, we have set EA parameters that are widely-used in the MOEA literature~\cite{trefzer2015evolvable}. In this set of experiments, all circuits are optimised with running $M=200$ generations with a population size $N$ of 200 individuals, and output load constraints have not been applied. We have run preliminary experiments with NSGA-II to confirm that reliably converges to similar performance when run multiple times with the same evaluation budget. Since the focus here is not on statistical analysis of the MOEA, we run the algorithm once for each benchmark to manage runtime. The number of synthesised gates and the number of genes are the same shown in Table~\ref{table:iscas_full} because all gates are encoded into chromosomes, so that the MOEDA flow is optimising the drive strength of all gates. In terms of the number of synthesised gates in each circuit, it is much less than the number in original benchmarks shown in Table~\ref{table:benchmarks}. It is the reason that the TSMC library has a large range of complex logic cells such as AOI (AND-OR-Inverter), IINR (NOR with 2 Inverted Inputs), full adders, etc., which are already comprised of few basic simple logic gates like XOR, NAND, OR, etc. In contrast, original benchmarks used basic simple generic gates. So this makes the synthesis tool to automatically merge the simple gates into complex ones for the total transistor count and physical area reduction, so finally reduce the number of gates.

This may compact the design space and reduce the search complexity but still increase the difficulty of PPA extra optimisation. In real-world libraries, complex logic cells have less options of drive strengths (normally no more than 5) due to the layout design complexity, and a large number of complex cells are used by tools evidenced by the significant decreasing in gate numbers. This may block the optimisation results for achieving huge improvements.

Under such difficulties, the MOEDA-optimised results are still promising compared to the Syn-Opt. solutions which are obtained by running the synthesis tool with ``try hard'' mode. MOEDA solutions demonstrate significant improvements in most test cases (up to 4.9\% in $D_{wc}$, 6.6\% in $P_{total}$ and 4.5\% in $A_{gate}$) as shown in Table~\ref{table:iscas_full}. The reported improvement of an objective does not (or slightly) sacrifice the metrics of other objectives. Although numbers might be small, even 1\% or 2\% of improvements could be helpful for designs under tight constraints and particularly when they just fail timing~\cite{fatemi2019enhancing}.

In Fig.~\ref{fig:MOO_Full}, the final generation of each circuit with three independent seeding runs is shown, plotting ``$D_{wc}$ vs. $P_{total}$'' (left column), ``$D_{wc}$ vs. $A_{gate}$'' (right column) and the corresponding Syn-Opt. reference solutions. The three clusters ($a$, $b$ and $c$) correspond to the three seed timing constraints, listed in Table~\ref{table:iscas_full}. In all cases, the MOEDA produces a wide range of useful trade-off solutions, with improved delay, reduced power consumption or area, within the boundaries of the given seed (Syn\_Opt. solution) topology.

From these plots, a number of solutions are improved regarding all objectives in four test circuits. The MOEDA-generated Pareto-driven clusters of C1908, C5315 and C6288 are smooth with good solution spreads, whereas the log2 circuit's is not. This is because the used algorithm in MOEDA flow needs to handle the increased size of design, where larger EA parameters (the number of generations $M$ and population size $N$) are required for producing Pareto-optimised results. The improved performance of log2 circuit is still promising and considerable in power and delay objectives under such an optimisation run with using the same EA parameters as other smaller test cases used. This implies that standard digital flow is also struggling to produce well trade-off solutions for a relatively larger design, so that the MOEDA flow has more optimisation room to get improved solutions run with relatively less iterations and a smaller population.

From smaller cases of C1908, C5315 and C6288, the tool's performance can be further observed when different constraints are applied. For timing settings corresponding to clusters $a$ and $b$, the tool is operating under tight timing requirements, causing the synthesis tool to spend the most effort on timing closure and less on power and area, so the MOEDA flow does not achieve significant improvements on delay (but much more trade-offs with less power and area). However, for relative relaxed timing settings corresponding to clusters $c$, the tool does not make the solution trade-off on timing too much but spend more efforts on power and area, where the MOEDA flow enhances the solution particularly in timing. This can conclude that the MOEDA flow demonstrates the capability of balancing these three objectives to a greater extend while tools have not.

Furthermore, as the circuit size increasing, the improvement of area is hard to be achieved (particularly in log2). This explicitly shows that area optimisation needs to include tuning the circuit structure (reducing gate count) instead of only focusing on drive strength refinement. But it is still worthwhile to take the area as one of objectives in the optimisation, which otherwise may have much degradation on area when optimising other objectives.

\subsection{Comparative Analysis with Stochastic Search}
\begin{figure}[t]
 \centering
    \subfloat{\includegraphics[width=0.5\linewidth]{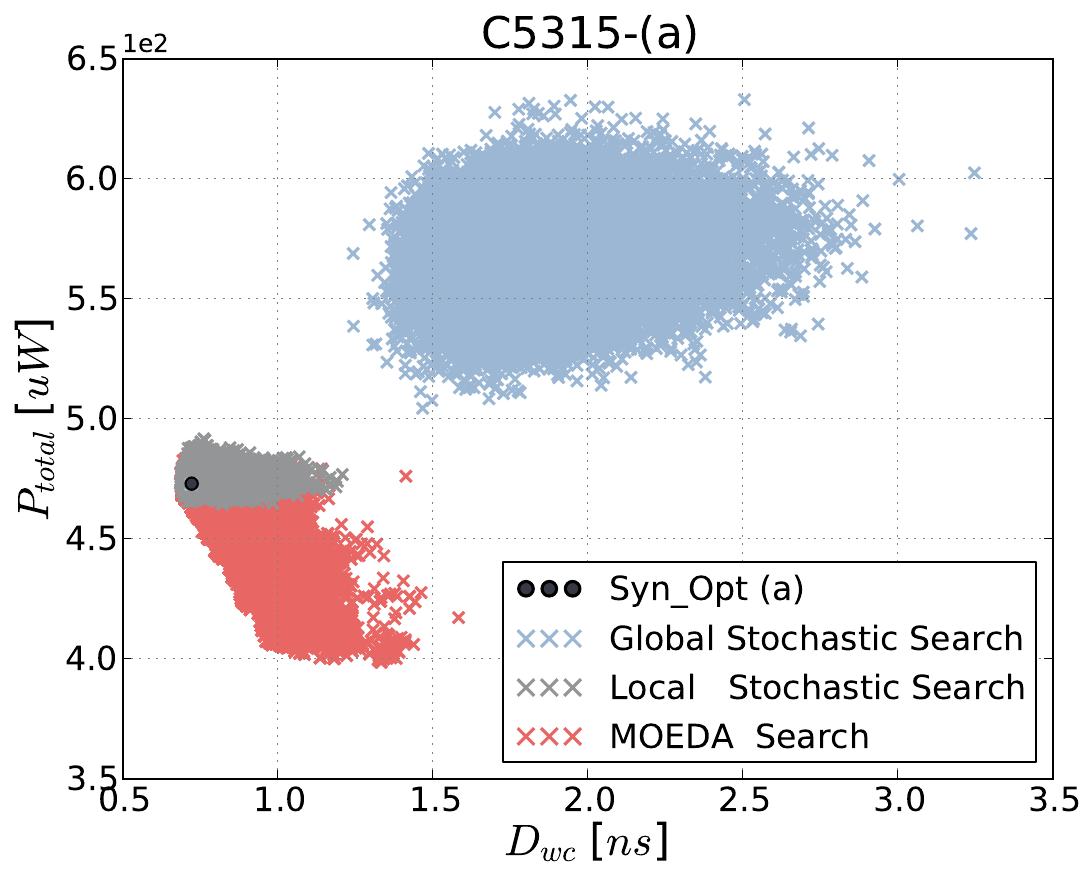}}
    \subfloat{\includegraphics[width=0.5\linewidth]{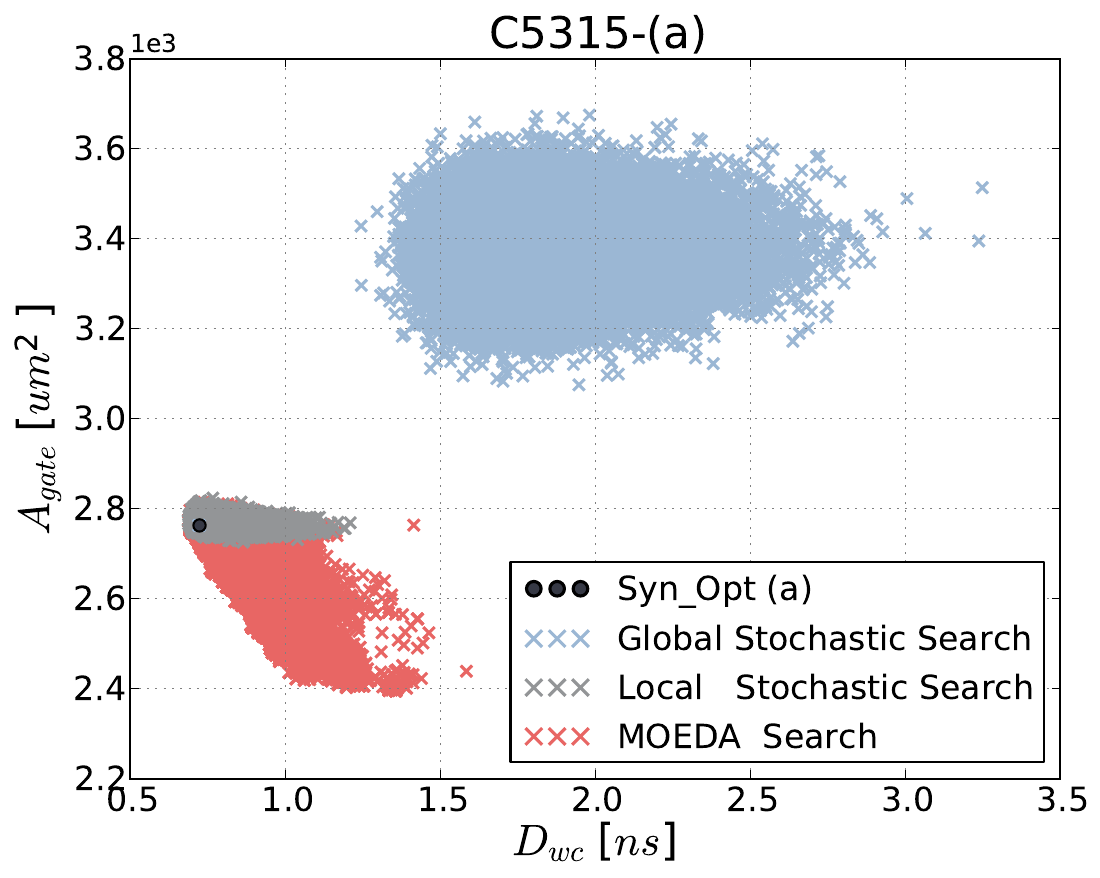}}
    \caption{MOEDA search compared to stochastic search, demonstrating the superior optimisation efficiency of evolutionary algorithms.}
    \label{fig:MOEDA_vs_SS}
\end{figure}

To demonstrate the optimisation efficiency of the MOEA used in the proposed MOEDA flow, we performed a comparative study between the MOEDA search and stochastic search. The selected test case is C5315-(a) from Section~\ref{section:exp:fulllib}, a 9-bit ALU with a tight timing constraint (a). For the MOEDA search, we initially run the NSGA-\RomanNumeralCaps{2} using a $1\%$ mutation rate with a 200-individual population size for 200 generations, so 40000 evaluations are generated in total. The MOEDA optimisation results (red cluster) shown in Fig.~\ref{fig:MOEDA_vs_SS} are seeded with the tool-optimised Syn\_Opt solution (black round scatter).

Two stochastic search experiments are then run here for comparison. The first one, referred to a local stochastic search (grey cluster shown in Fig.~\ref{fig:MOEDA_vs_SS}), is to randomly produce 40000 individuals seeding with the same Syn\_Opt solution, and each of them is achieved by randomly mutating the chromosome using the same probability (1\%). The second one is completely randomised results referred to a global stochastic search, which produces 40000 individuals seeding with the same Syn\_Opt solution but all genes (i.e., drive strengths of logic gates) of each individual are modified (100\% mutation rate). The results of global stochastic search are the blue cluster shown in Fig.~\ref{fig:MOEDA_vs_SS}.

Based on the observations made from these plots, it demonstrates that the NSGA-\RomanNumeralCaps{2} algorithm used in the MOEDA flow has superior optimisation performance when compared to the stochastic search. Since the focus of this work is not on investigating which MOEA is the best to achieve the optimum results in VLSI design optimisation, only NSGA-\RomanNumeralCaps{2} is used here for experiments.

\subsection{Discussion}
The runtime for the largest case optimisation (log2.a) needs 138 hours. Although the proposed optimisation method is at the cost of longer computing time, this investment will be worthwhile when considering the enhancements in delay and savings in power consumption or area that could not otherwise be achieved, particularly for feasible circuit solutions that are produced in large numbers.

In addition, optimising circuits for a given timing constraint with one circuit topology solution (single seed) is not capable enough to offer a larger design space when circuit structures are changing. Therefore, the next section will investigate how running synthesis multiple times can be harnessed to expand, access and explore the design space with respect to different circuit topologies.

\section{Multi-objective Design Space Exploration}
\label{section:DSE}

\begin{table}[t]
\begin{center}
\caption{Timing Constraints of Each Benchmark}
\begin{tabular}{|c|c|c|c|c|}
  \hline
  Test  & clock     &$T_{r}$            &set              &\# Syn Gates \\ 
  Case  & ($T_{c}$) &(Increment Factor) &load             &\# Genes     \\\hline
  \multirow{2}{*}{C1908}     &250MHz  &$1.50ns-0.51ns$   &D1 &105 - 445     \\\cline{4-5}
                             &($4ns$) &($0.01ns$)        &D8 &105 - 468     \\\hline
  \multirow{2}{*}{C5315}     &250MHz  &$1.50ns-0.51ns$   &D1 &396 - 1323    \\\cline{4-5}
                             &($4ns$) &($0.01ns$)        &D8 &401 - 1287    \\\hline
  \multirow{2}{*}{C6288}     &250MHz  &$4.00ns-2.02ns$   &D1 &1105 - 3208   \\\cline{4-5}
                             &($4ns$) &($0.02ns$)        &D8 &1123 - 3222   \\\hline
  \multirow{2}{*}{log2}      &40MHz   &$25.00ns-15.10ns$ &D1 &10801 - 12561 \\\cline{4-5}
                             &($25ns$)&($0.10ns$)        &D8 &10797 - 12555 \\\hline
\end{tabular}
\label{table:design_space_summary}
\end{center}
\end{table}

\captionsetup[subfigure]{labelformat=empty}
\begin{figure*}[!ht]
  \centering
  \subfloat{\includegraphics[width=0.245\linewidth]{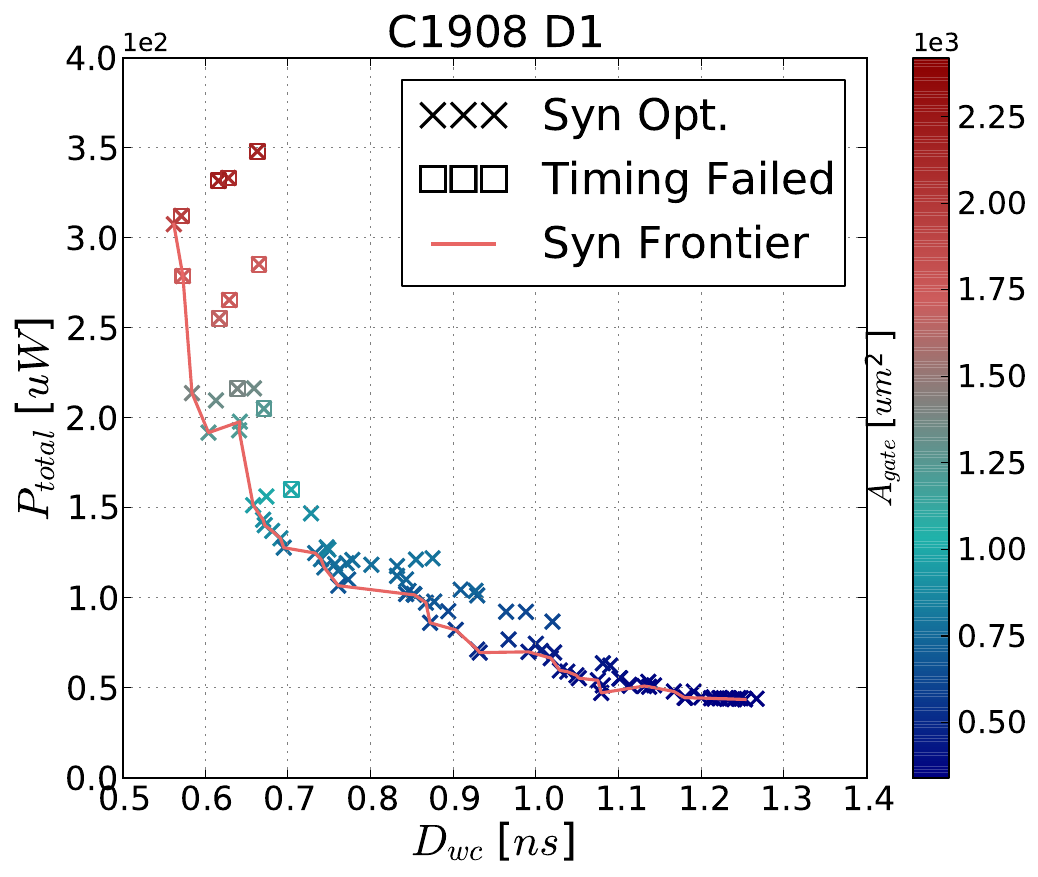}}
  \subfloat{\includegraphics[width=0.255\linewidth]{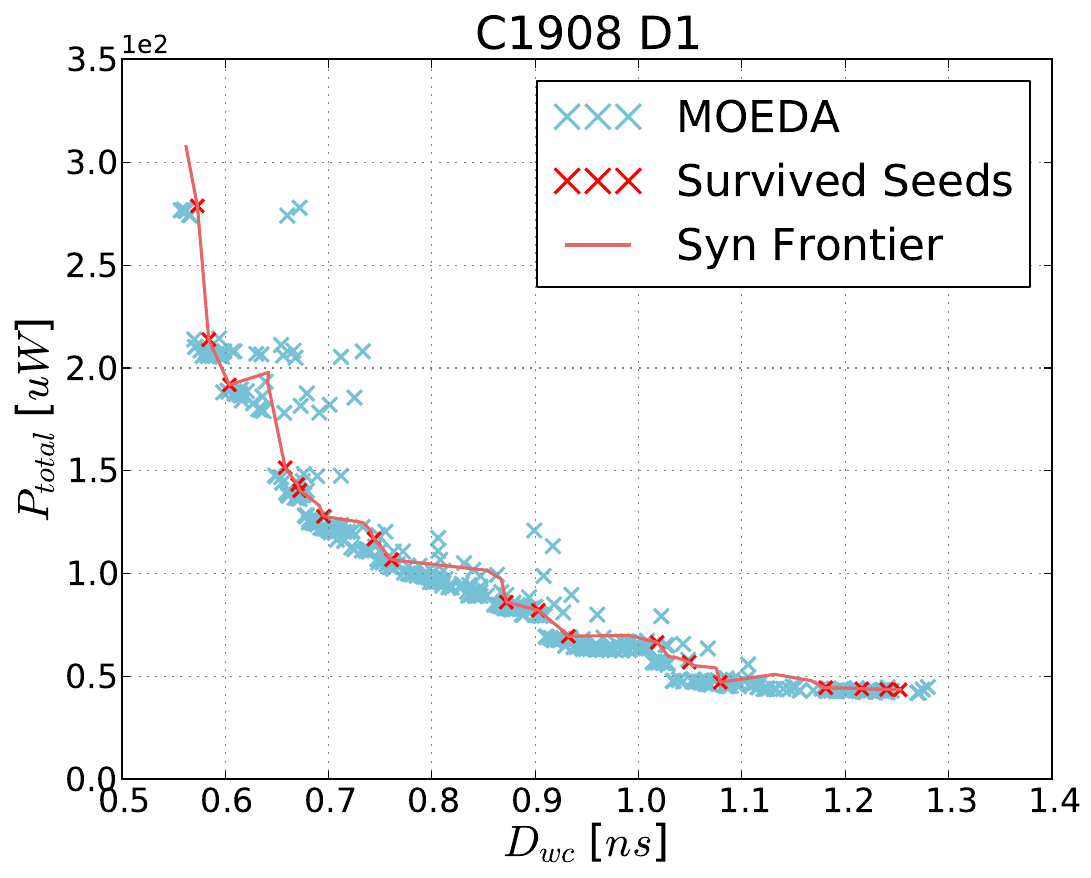}}
  \subfloat{\includegraphics[width=0.245\linewidth]{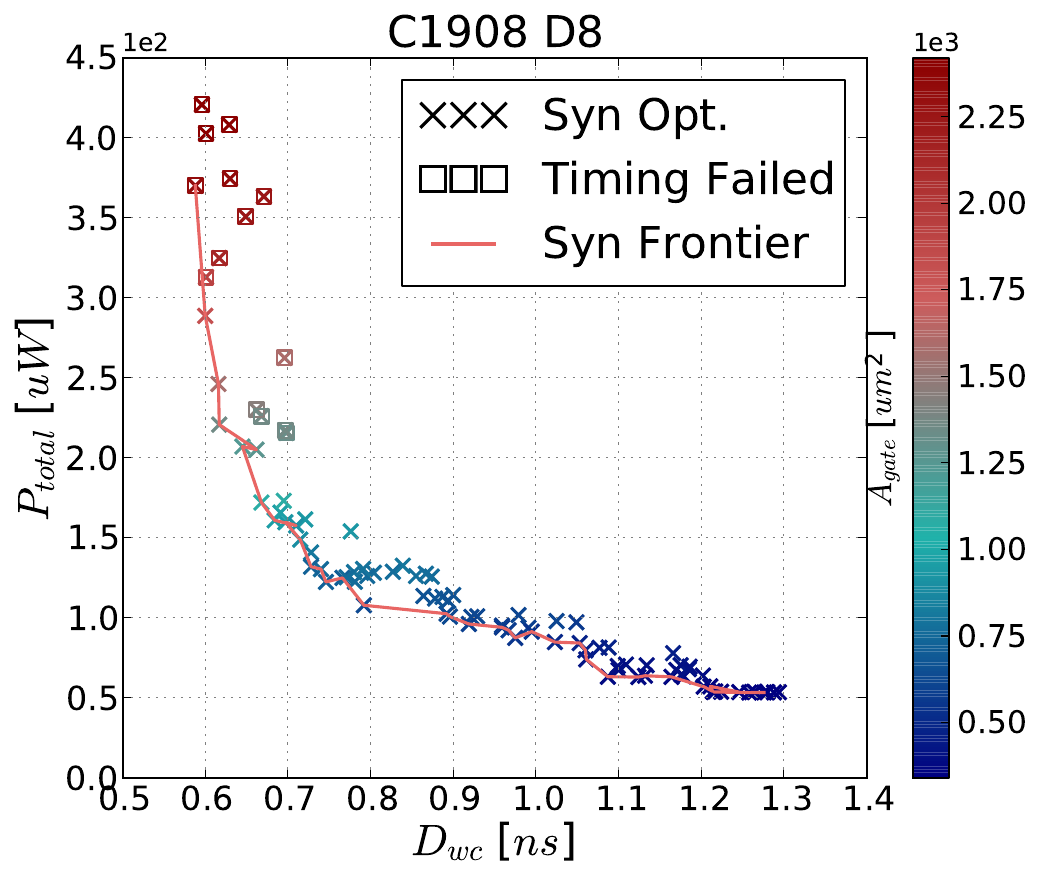}}
  \subfloat{\includegraphics[width=0.255\linewidth]{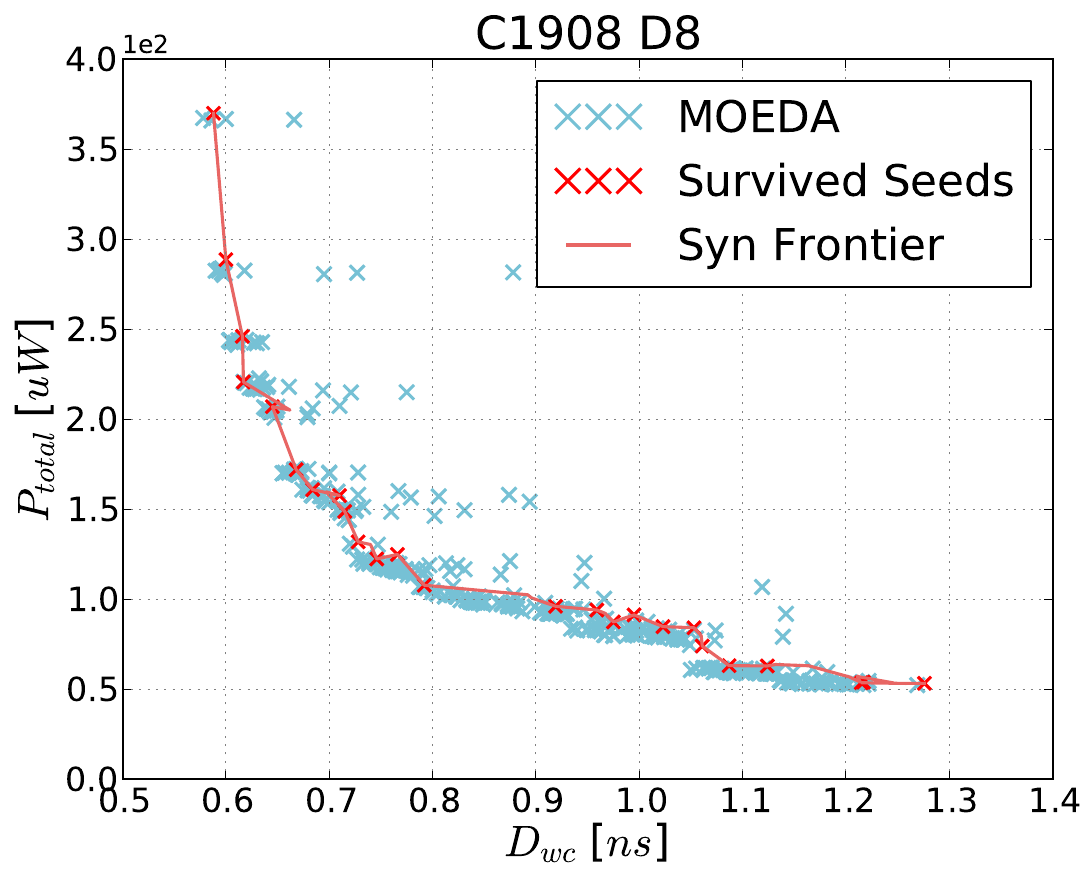}}\\
  \subfloat{\includegraphics[width=0.245\linewidth]{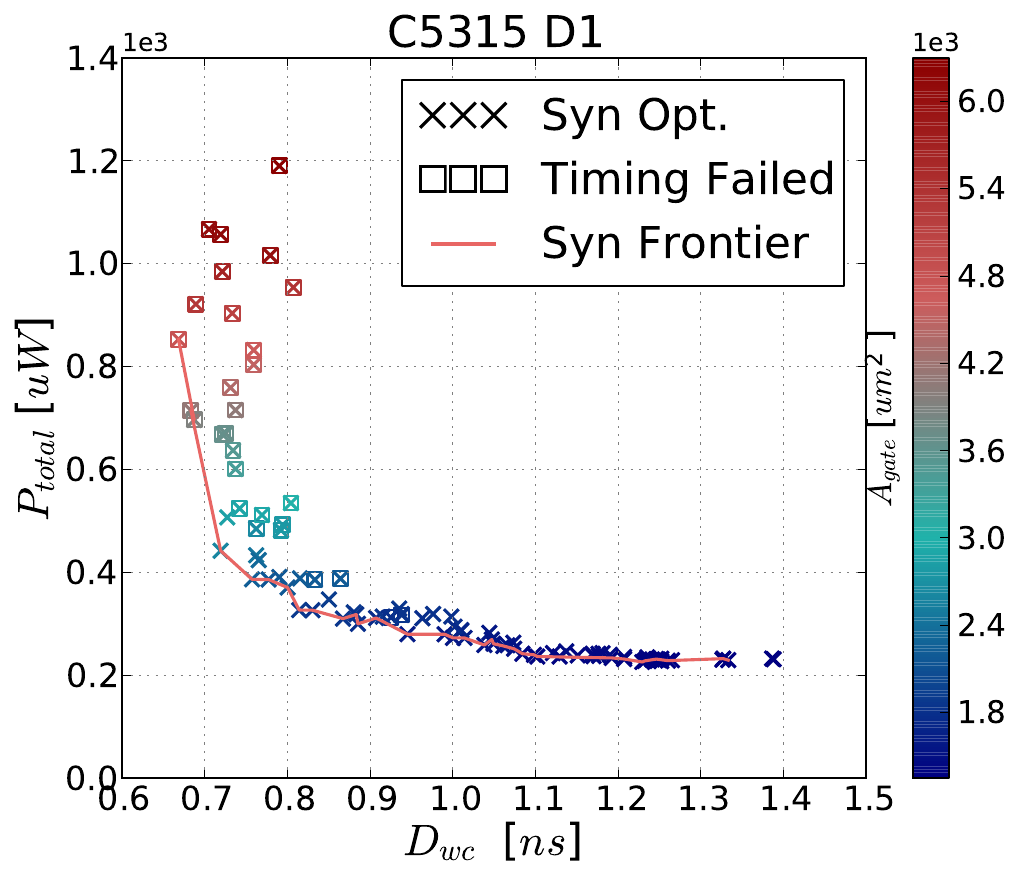}}
  \subfloat{\includegraphics[width=0.255\linewidth]{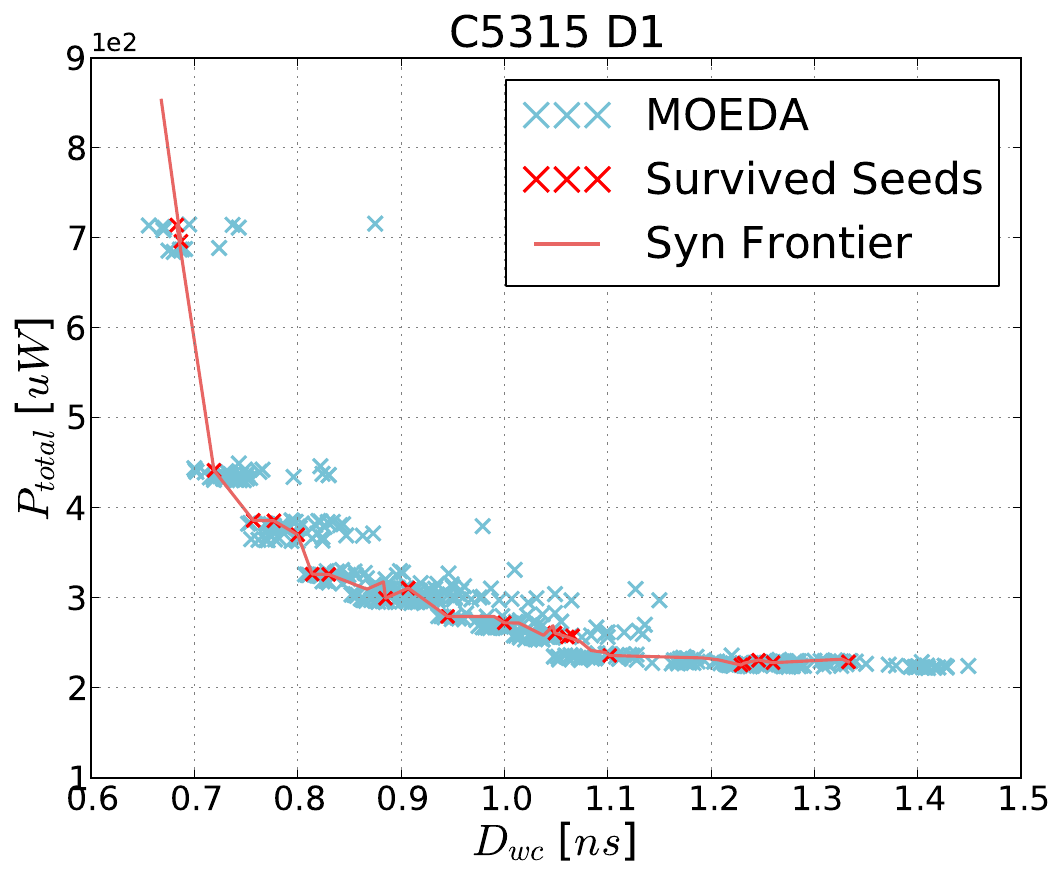}}
  \subfloat{\includegraphics[width=0.245\linewidth]{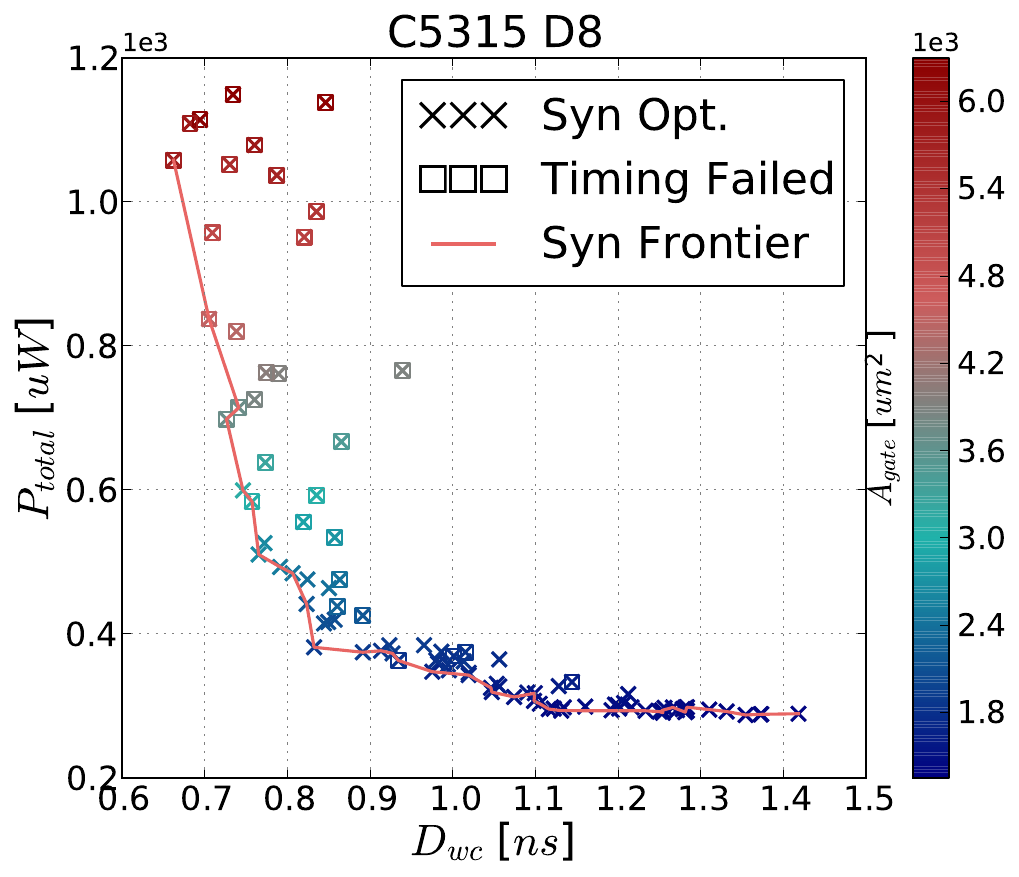}}
  \subfloat{\includegraphics[width=0.255\linewidth]{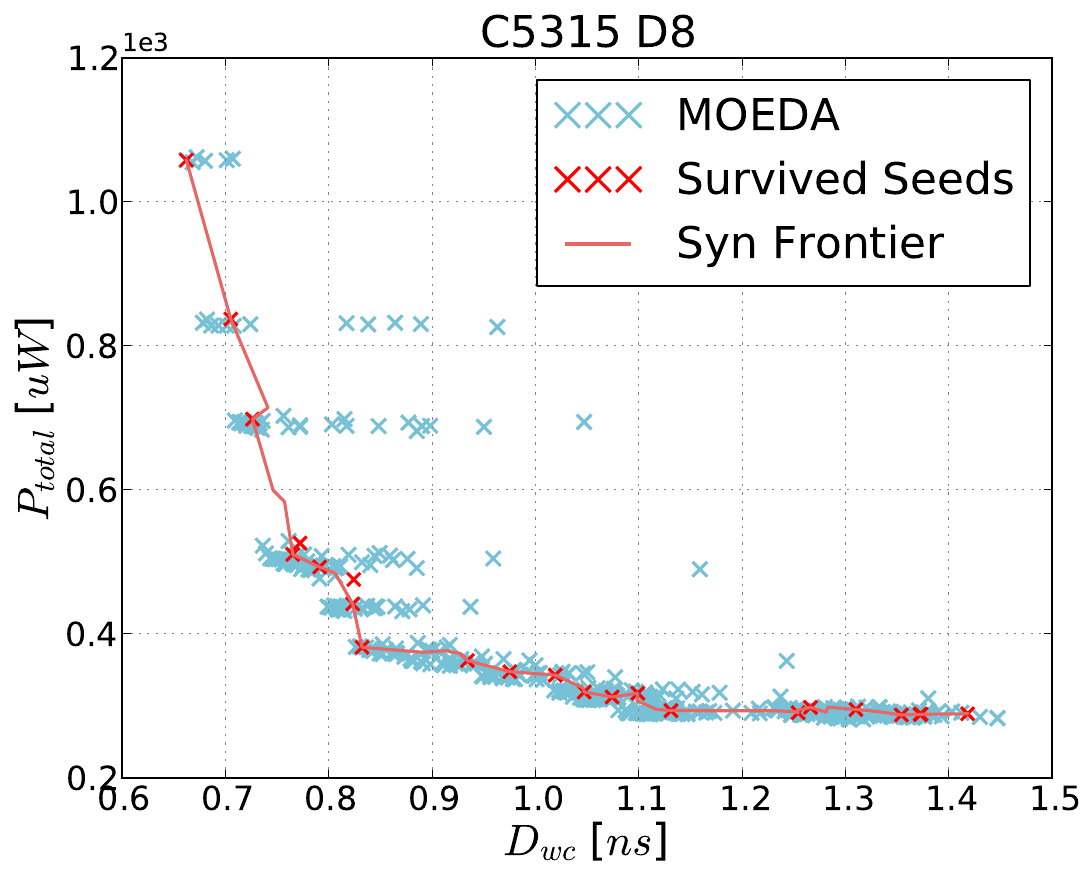}}\\
  \subfloat{\includegraphics[width=0.245\linewidth]{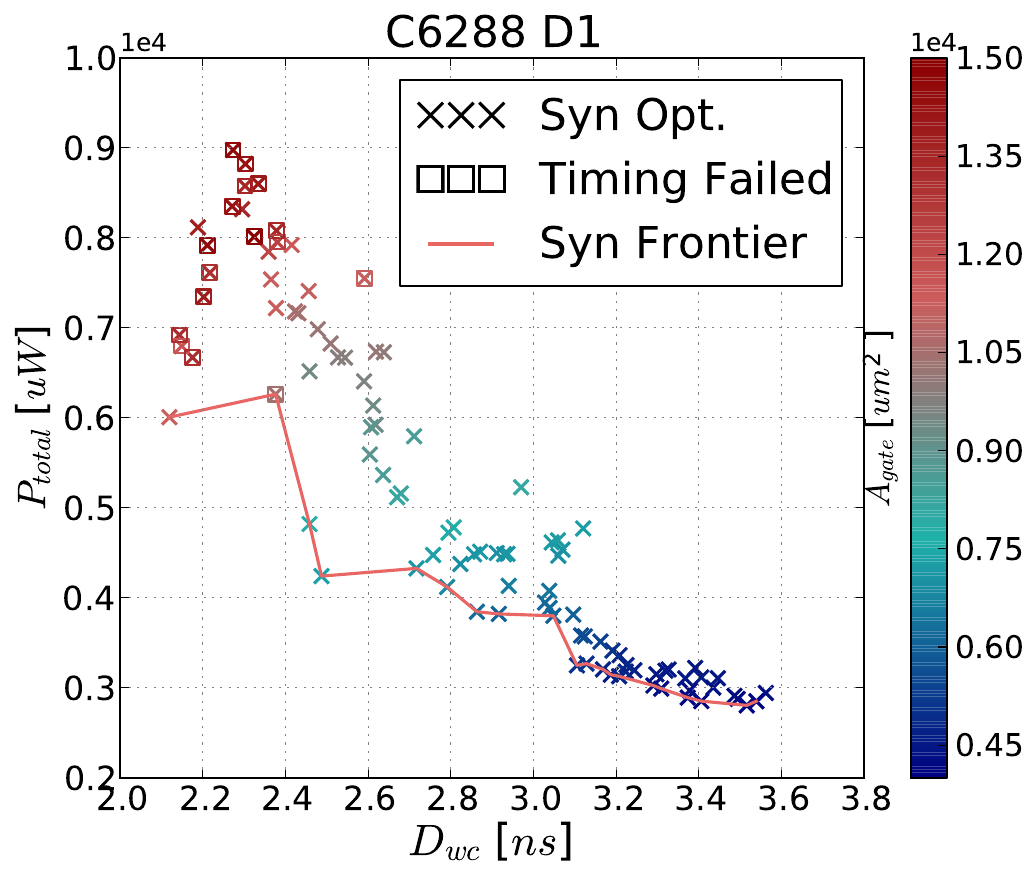}}
  \subfloat{\includegraphics[width=0.255\linewidth]{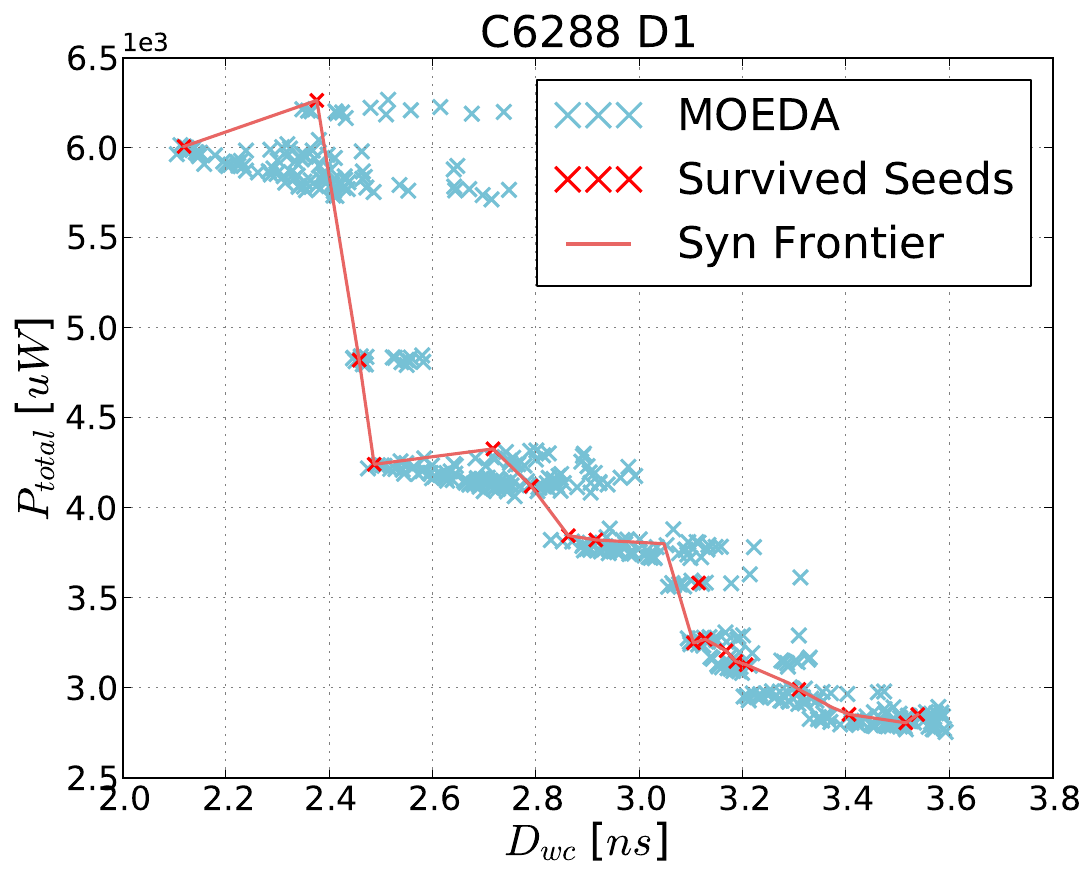}}
  \subfloat{\includegraphics[width=0.245\linewidth]{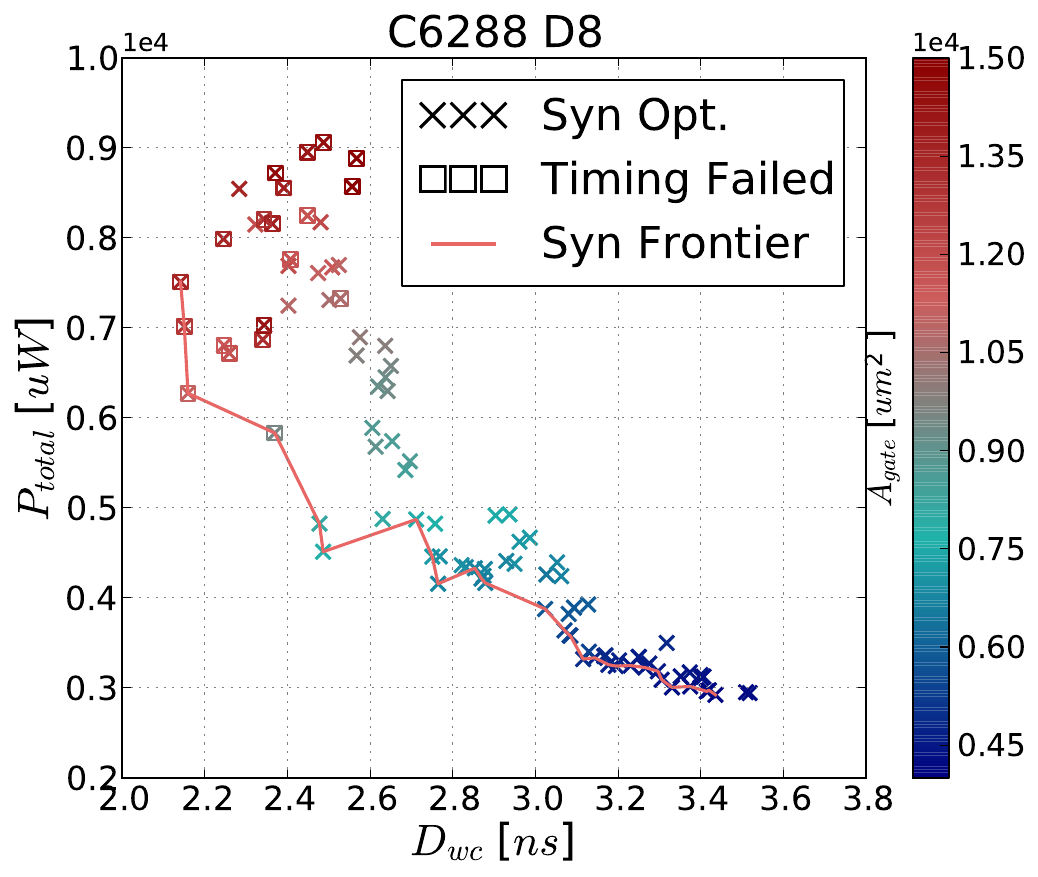}}
  \subfloat{\includegraphics[width=0.255\linewidth]{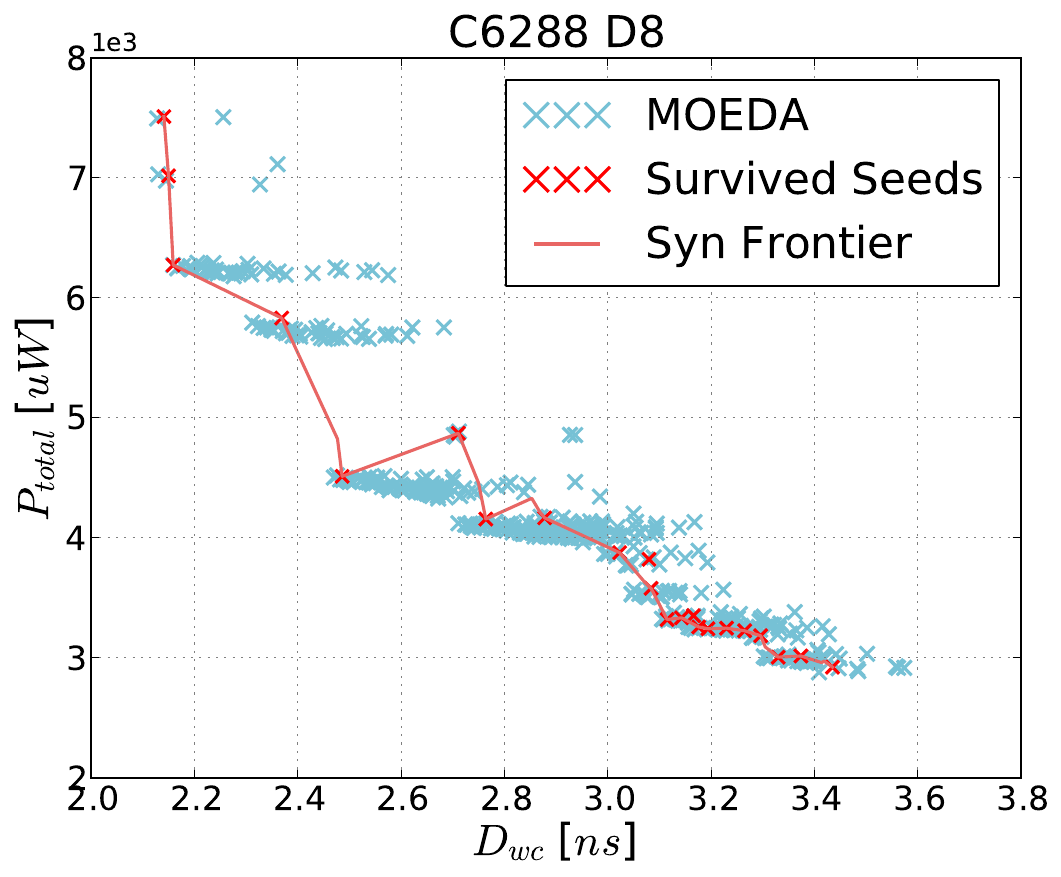}}\\
  \subfloat[Standard Flow]{\includegraphics[width=0.235\linewidth]{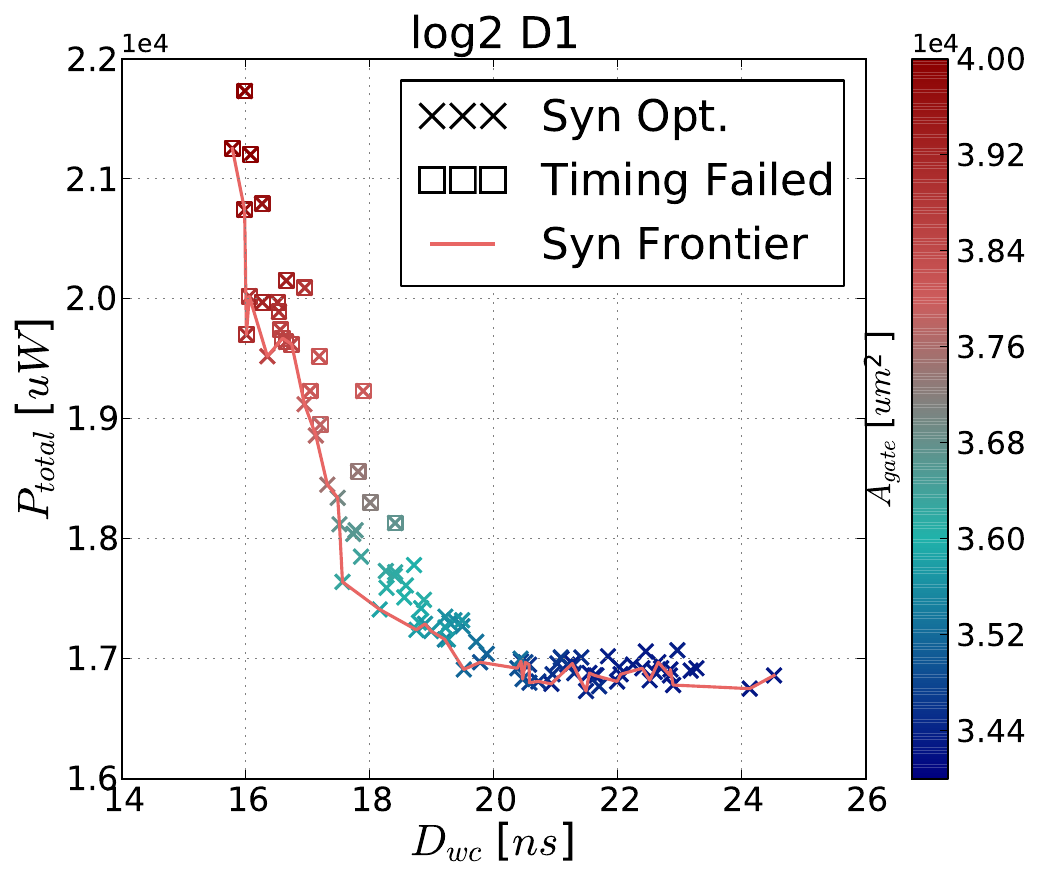}}
  \subfloat[MOEDA Flow]{\includegraphics[width=0.265\linewidth]{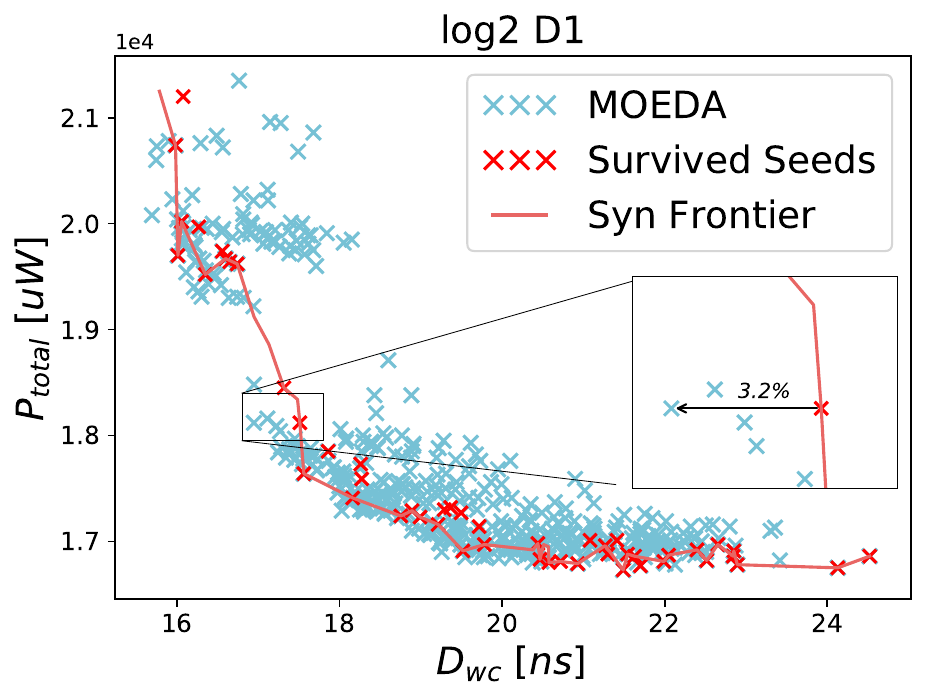}}
  \subfloat[Standard Flow]{\includegraphics[width=0.235\linewidth]{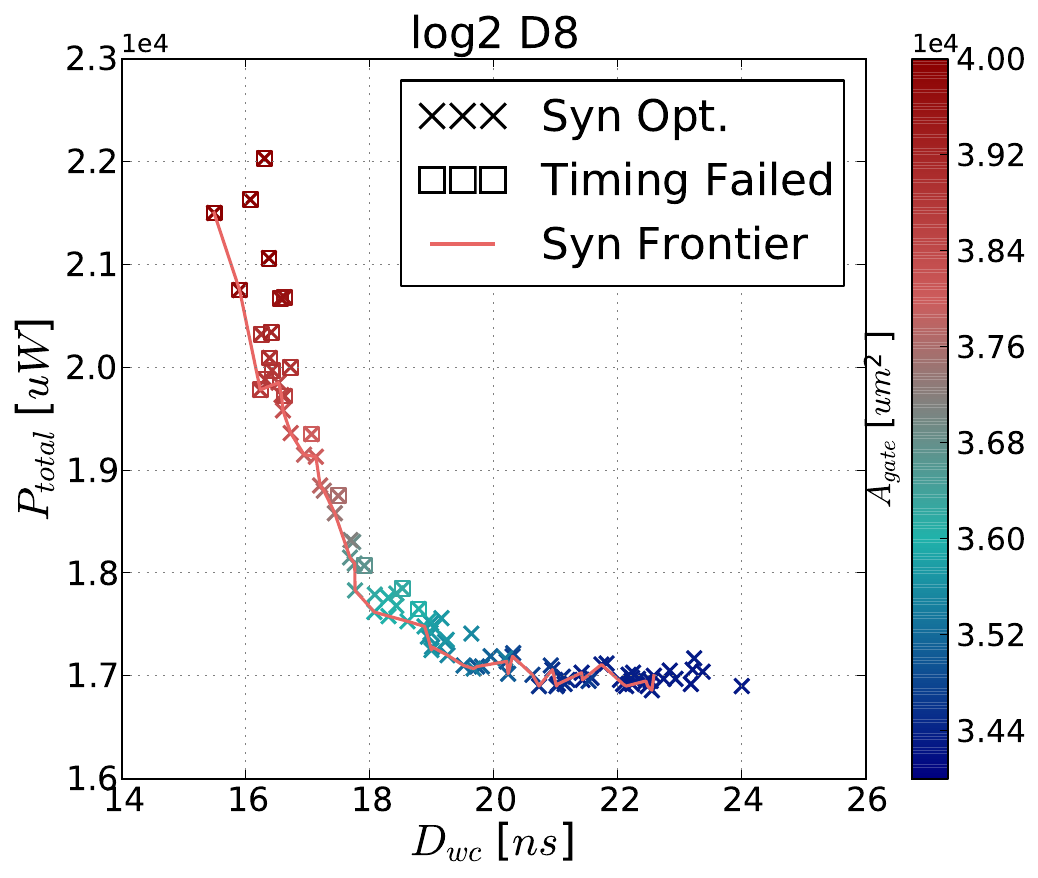}}
  \subfloat[MOEDA Flow]{\includegraphics[width=0.265\linewidth]{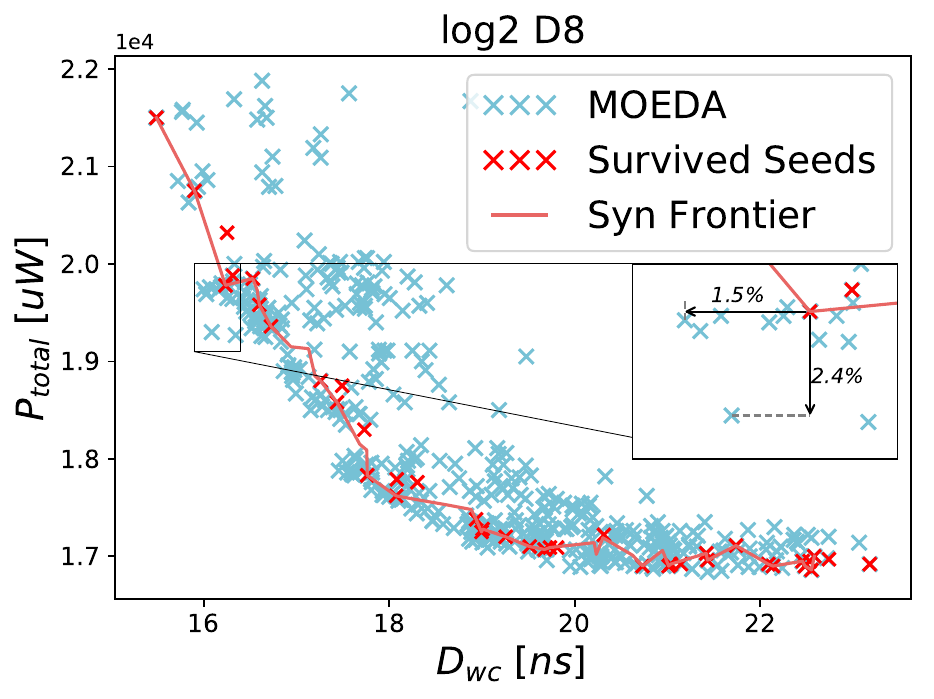}}
  \caption{Design space optimisation results under the drive strength D1 and D8 output load scenarios for C1908 16-bit error detector/corrector, C5315 9-bit ALU, C6288 16x16 multiplier and log2 calculation circuit. $N=500$, $M=100$, $\rho=1\%$, set\_load information is labeled in the title at the top of each plot. The optimised design space of log2 with D1 and D8 loads is shown with zoom-in views to present the improvements clearly.}
  \label{fig:MOO_DSE}
\end{figure*}

\subsection{Optimisation using Multiple Seed Designs}
Instead of seeding the initial population using a single synthesis-optimised solution for a separate MOEDA run, this section investigates how the proposed algorithm can explore the design space simultaneously using a set of multiple different seeds. The seeds are a range of different solutions generated using the standard digital flow under a number of different timing constraints.

The methodology to obtain different seeds from the standard design tools is the same as before in Section~\ref{section:exp:fulllib}. However, in this case, a more fine-grained range of timing constraints are applied in 100 increments from minimum (a constraint that the tool can easily meet) to maximum (solutions start to fail timing) in order to investigate what design space coverage they can achieve. Each benchmark has been synthesised once for each timing constraint setting to generate the 100 solutions for seeding. Table~\ref{table:design_space_summary} summarises timing constraint settings of each test case including the number of synthesised gate from minimum to maximum. Different output load scenarios, including loading with drive strength D1 and D8, are applied to the outputs of all test cases under the same set of timing constraints. The output load values (D1 and D8) are specified as the input pin capacitance of inverters with drive strength D1 and D8 from the TSMC cell library. The reason of selecting D1 and D8 as output loads is that D1 load is a nominal scenario in practice and D8 load with larger capacitance is the middle sized one from all available inverters.

The first and third columns of Fig.~\ref{fig:MOO_DSE} illustrate the standard tool's design space for each benchmark circuit under D1 and D8 output load scenarios. Their respective optimised design space from the MOEDA flow is shown in Fig.~\ref{fig:MOO_DSE}'s second and fourth columns. From ``Standard Flow'' columns, all cross markers represent tool-generated solutions in ``$D_{wc}$ vs. $P_{total}$'' and their face colors correspond to the color bar relating to the area objective $A_{gate}$ ranging from large (red) to small (blue). Solutions additionally marked with squares have failed to meet timing constraints. The red line highlights the Syn-Opt.~\enquote{elite} solution front, which is calculated using the non-dominated sorting approach in three dimensions in regard to $D_{wc}$, $P_{total}$ and $A_{gate}$. All solutions in the first domination rank are connected with a line to highlight the~\enquote{Syn-Frontier} more clearly. The Syn-Frontiers shown in the figures are projections from the 3D objective space onto the 2D plots.

Looking at the design space of the standard flow, it can be observed that the 16-bit error detector/corrector (C1908), the 9-bit ALU (C5315) and even the log2 circuit can be synthesised and optimised well by the tool as the set of solutions forms a smooth Pareto frontiers. However, the 16x16 multiplier (C6288), which is a highly structured circuit using a number of adders, yields a less regular frontier with more clustered solutions. This indicates that the synthesis tool struggles to effectively trade-off multiple objectives when optimising a complex design with a relatively fixed circuit topology.

\subsection{Squeeze Design Space for PPA}
The design space comprising the 100 seed solutions, in different circuit topologies, is the baseline for the MOEDA to perform optimisation on. All 100 seed solutions are loaded into the initial population of the MOEDA flow and optimised generation by generation. All test cases are optimised over 100 generations using a population size of 500, i.e. the initial population comprises five copies of each seed circuit. The plots in ``MOEDA Flow'' columns of Fig.~\ref{fig:MOO_DSE} show the improved solution space, plotting ``$D_{wc}$ vs. $P_{total}$''. The red line shows the Syn-Frontiers from the baseline design space. All those seed solutions that have survived until the last generation, although with modified drive strengths, are marked with a red cross. The solutions shown as blue crosses are those produced by the MOEDA flow comprising of all individuals of the final generation.


The results confirm that the MOEDA flow can push the baseline frontier further to extend the design space of all test cases in all three objectives, through different circuit topologies. For the largest circuit log2, the optimised design space is shown with additional zoom-in views to present the quantified improvements which are still considerable. Furthermore, the relative improvement looks marginal from the plots due to the wide axis range, but the total absolute values for saved power and improved delay are significant.

In the case of circuit C1908, the optimised solutions that form a smooth Pareto frontier, whereas there are some gaps in the optimised design space of C5315, log2 and particularly of C6288. The gaps are artefacts from the baseline design space due to limitations of the tool's optimiser and properties of the circuit. Although the proposed MOEDA flow could not fully bridge these large gaps, it has been achieved that the optimised design space covers the baseline design space and beyond more uniformly. This makes better choices for design-specific using as a richer set of solutions is available.

Only about one-fourth of the initial seeds survive until the final generation in design C1908, C5315 and C6288, and about half of the initial seeds survive in log2 circuit. Most of the surviving seeds are positioned on the Syn-Frontier, while others have been discarded in the evolution process. This indicates that there is ``noisiness'' inside of standard flows/tools and not all solutions generated by tools are presumably optimised, which might lose some well trade-offed solutions. This normally requires iterations with applying modifications in the design flow achieved by engineers with custom design efforts. The MOEDA can auto-iterate designs without throughout the whole flow for better trade-offs in PPA metrics.

\begin{table}[t]
\begin{center}
\caption{Search Efficiency of MOEDA}
\begin{tabular}{|c|c|c|c|c|}
  \hline
  Test                     &set  &No. Pareto Solutions &Total No.                &Search\\ 
  Case                     &load &in Final Generation.        &Evaluations              &Efficiency (\%)\\\hline
  \multirow{2}{*}{C1908}   &D1   &304        &\multirow{2}{*}{50,000}  &0.61\% \\
                           &D8   &307        &                         &0.61\% \\\cline{1-5}
  \multirow{2}{*}{C5315}   &D1   &254        &\multirow{2}{*}{50,000}  &0.51\% \\
                           &D8   &250        &                         &0.50\% \\\cline{1-5}
  \multirow{2}{*}{C6288}   &D1   &257        &\multirow{2}{*}{50,000}  &0.51\% \\
                           &D8   &259        &                         &0.52\% \\\cline{1-5}
  \multirow{2}{*}{log2}    &D1   &150        &\multirow{2}{*}{50,000}  &0.30\% \\
                           &D8   &166        &                         &0.33\% \\
  \hline
\end{tabular}
\label{table:search_efficiency}
\end{center}
\end{table}

To demonstrate the search efficiency of the MOEDA, Fig.~\ref{fig:MOO_DSE} also  includes the MOEDA-optimised non-dominated solutions covering all three objectives to show the relative position of the optimised and the initial tool-generated ones. The results clearly show that the optimised front dominates the initial tool-generated one. Table~\ref{table:search_efficiency} further summarises the quantified quality of the Pareto solution sets compared to the total number of designs explored. The No. of Pareto solutions presented in the table is the total non-dominated solutions of the final MOEDA generation in each test case. The total number of evaluations is 50,000 which is the same for all cases. The search efficiency is then obtained by calculating the ratio of the Pareto solutions in the final generation to the total number of evaluations. In log2 circuit, the efficiency is slightly lower than for other designs due to its larger size.

\subsection{Discussion}
The runtime of largest case (log2.D8) is 162 hours. The MOEDA flow needs more computing resources due to the continuous generation of design layouts. This aims for accurate and real-world evaluation. It is easily to speed up the flow through making design evaluations at earlier design stage without place and route, but what we are investigating in this work is whether the proposed MOEDA flow has generic optimisation capability in an industrial environment, and the MOEDA flow has feasibly improved the performance of block circuit instances used in this work. With regard to scalability, in terms of design size, an iterative critical path optimisation for extreme-large designs (e.g., millions of gates) using MOEDA flow is also our work in progress, with the potential aim to solve timing violations faster and still without increasing the power or area.

The MOEDA flow achieves significant improvements on PPA over the standard design tool's solutions across the entire design space with different circuit topologies. However, although the proposed method is capable of exploiting design opportunities to refine technology mapping by adjusting drive strengths at the gate-level, circuit topology optimisation is currently not yet included. This current limitation is likely the reason that design space gaps cannot be fully closed, which would provide the best trade-off design choices. This is particularly visible in the results for C6288, due to its fixed topology. From these results it can be envisaged that including topology modification in our approach could enable further design optimisation opportunities, particularly in the case of complex circuits with rigid structure.

\section{Conclusions}
\label{section:Conclusion}
This paper proposes a fully-automated multi-objective electronic design automation flow (MOEDA) extension to enhance the current industry-standard synthesis and physical implementation flow, primarily suited for IP/block level designs. The MOEDA flow is fully compatible with commercial design tools and specifically optimises drive strength of gates during technology mapping in such a way that the subsequent physical implementation stage can achieve designs with better PPA metrics. The proposed method has been successfully applied to the optimisation of designs from ISCAS-85 and EPFL benchmark suite using the TSMC 65nm low power standard cell library. 

Experimental results show that the proposed MOEDA flow has operated design optimisation gaining significant improvements on PPA over the standard tool's solutions. It can be concluded that optimising technology mapping to refine drive strength selection of cells is beneficial to improving PPA of circuits. This has not only been shown for a single solution, but across the entire design space with various circuit topologies.

From a designer's point of view, the multi-objective optimisation approach has the added benefit of producing a set of best trade-off solutions which are as uniformly as possible distributed. This provides designers with choice and allows to select designs with the most appropriate objective trade-off for different applications.

\bibliographystyle{IEEEtran.bst}
\bibliography{VLSI}



\end{document}